\documentclass{egpubl}
\usepackage{eg2019}

\ConferencePaper        
\electronicVersion 

\ifpdf \usepackage[pdftex]{graphicx} \pdfcompresslevel=9
\else \usepackage[dvips]{graphicx} \fi

\PrintedOrElectronic

\usepackage{t1enc,dfadobe}
\usepackage{egweblnk}
\usepackage{cite}
\usepackage{flushend}
\usepackage{amsmath}
\usepackage{amssymb}

\usepackage{booktabs}
\usepackage{graphicx}
\usepackage[export]{adjustbox}
\usepackage{multirow}
\usepackage{xcolor}
\usepackage{float}

\DeclareMathOperator*{\argmax}{arg\,max}
\newcommand{\h}{0.363\linewidth}
\newcommand{\hh}{0.0985\linewidth}

\usepackage{bookmark}
\bookmarksetup{
  numbered,
  open,
}

\title[Learning to Importance Sample in Primary Sample Space]%
      {Learning to Importance Sample in Primary Sample Space}

\author[Q. Zheng \& M. Zwicker]
{\parbox{\textwidth}{\centering Quan Zheng\orcid{0000-0001-5053-5511} and 
Matthias Zwicker\orcid{0000-0001-8630-5515}
                    }\\
{\parbox{\textwidth}{\centering University of Maryland, College Park, USA\\
                    quan.zheng@outlook.com; zwicker@cs.umd.edu
                    }
}
}

\begin{document}


\maketitle
\begin{abstract}
   Importance sampling is one of the most widely used variance reduction strategies in Monte Carlo rendering. We propose a novel importance sampling technique that uses a neural network to learn how to sample from a desired density represented by a set of samples. Our approach considers an existing Monte Carlo rendering algorithm as a black box. During a scene-dependent training phase, we learn to generate samples with a desired density in the primary sample space of the renderer using maximum likelihood estimation. We leverage a recent neural network architecture that was designed to represent real-valued non-volume preserving (``Real NVP'') transformations in high dimensional spaces. We use Real NVP to non-linearly warp primary sample space and obtain desired densities. In addition, Real NVP efficiently computes the determinant of the Jacobian of the warp, which is required to implement the change of integration variables implied by the warp. A main advantage of our approach is that it is agnostic of underlying light transport effects, and can be combined with an existing rendering technique by treating it as a black box. We show that our approach leads to effective variance reduction in several practical scenarios. \\

\end{abstract}  
\section{Introduction} \label{sec:intro}

Importance sampling has been recognized as a key technique for variance reduction right from the inception of Monte Carlo rendering algorithms~\cite{kajiya1986rendering}. Today, importance sampling of BRDFs, environment maps, direct illumination from many light sources, or visibility are standard features in Monte Carlo path tracing systems. A number of advanced techniques have also been developed to jointly importance sample several of these factors. Many of these approaches rely on an analytical analysis of scene properties, such as the surface appearance models and BRDFs used in the scenes.

In this paper, we propose a technique that treats an existing Monte Carlo renderer as a black box and learns how to importance sample entire paths in primary sample space. Our approach first acquires a set of training samples for a given scene using the existing renderer. Based on these samples, we learn to generate a desired scene-dependent target density in the primary sample space (PSS) of the renderer. In the subsequent rendering step, instead of feeding the renderer with uniform PSS samples, we provide samples drawn from the learned target density. By specifying a suitable target density, we achieve effective variance reduction compared to using the existing renderer with the usual uniform primary sample space.

The key component of our approach is a recent neural network architecture that was designed to represent real-valued non-volume preserving (``Real NVP'') transformations in high dimensional spaces. This approach learns a one-to-one, non-linear warp between two high-dimensional spaces. In addition, the computation of the warp is structured such that the forward warp, its inverse, and the determinant of its Jacobian can all be computed effectively. We leverage these properties to learn a warp from a uniform density to a desired non-uniform target density in primary sample space, and then generate well-distributed PSS samples for Monte Carlo rendering tasks. {Accounting for efficiency issues with high-dimensional spaces, we adopt practical simplifications to make learning and sampling the target density tractable in limited dimensions}. The advantages of our approach are that it treats a renderer as a black box and is agnostic to specific light transport effects, hence it can be combined with many existing algorithms.

Conceptually, our approach has similarities to several previous strategies. Primary sample space Metropolis (PSS-MLT) light transport can also importance sample any desired target density by operating in primary sample space while treating an existing renderer as a black box~\cite{kelemen2002}. PSS-MLT can be inefficient, however, because it often needs to reject many proposed paths to achieve the desired density. In contrast, our approach never rejects samples during rendering. Several previous techniques acquire a set of initial samples to build up low-dimensional ad hoc data structures that approximate the desired density, and then can be used to perform importance sampling during subsequent rendering passes. The advantage of our approach is that the renderer is treated as a black box, and the approach does not rely on additional data structures such as octrees, spatial samplings of the scene, or explicit density models. Instead, the information learned from an initial set of samples is captured by the Real NVP neural network. In summary, the main contribtions of this paper are:
\begin{itemize}
    \item A novel formulation of importance sampling of entire light paths as a non-linear warp in primary sample space.
    \item A novel technique to learn the primary sample space warp using a suitable neural network architecture.
    \item A demonstration that this approach can effectively reduce variance of Monte Carlo path tracers in several scenarios.
\end{itemize}

\section{Related Work} \label{sec:relatedwork}

\subsection{Importance Sampling in Monte Carlo Rendering}\label{sec:importancesampling}

Already when introducing the rendering equation, Kajiya~\shortcite{kajiya1986rendering} discussed importance sampling as a technique to reduce variance in Monte Carlo rendering. Importance sampling aims to obtain samples with a probability density that is proportional to a desired target density function, and by designing target densities similar to the integrand, variance can be reduced. In Monte Carlo path tracing, the target density function can be defined either as a product of local densities in an incremental manner, or directly as a density in global path space. Accordingly, two categories of importance sampling techniques have been studied in past decades.

\textbf{Incremental sampling.} \,When light paths are constructed in an incremental way, it is natural to perform importance sampling locally in each step. For example, BRDF sampling and light sampling techniques construct a path segment by separately mimicking the local distribution of cosine weighted BRDF and incident illumination, respectively. To further reduce the estimator's variance, Veach and Guibas~\shortcite{veach1995} introduced multiple importance sampling to combine the advantages of individual path sampling techniques. We refer to standard texts for an overview of the extensive literature~\shortcite{pbrt}. In general, importance sampling techniques can be categorized into two broad groups: \emph{a priori} methods that construct target densities and sampling techniques using analytical approximations of the integrand (for example, Heitz and d'Eon~\shortcite{Heitz2014ISM} among many others), and \emph{a posteriori} methods that fit target densities based on empirically acquired samples of the integrand~\cite{vorba2014line,muller2017practical, Dahm17}. The technique by Clarberg~\shortcite{Clarberg:2005:WIS} is an \emph{a priori} method that performs importance sampling of products of environment maps, BRDFs, and visibility by representing them using wavelets. They achieve importance sampling by evaluating wavelet products on the fly and constructing a hierarchical warp accordingly, which has some similarities to our warp-based importance sampling. Our approach, however, operates in an \emph{a posteriori} manner and it does not rely on a specific representation of the integrand, such as using wavelet products. Our approach is also more general and can be applied to any light transport effect. Recently, a Bayesian method has been proposed to sample direct illumination that can be considered a combination of \emph{a priori} and \emph{a posteriori} strategies~\cite{Vevoda2018:BOR}. Our approach considers entire light paths instead of performing incremental sampling.

\textbf{Global sampling.} \,Instead of importance sampling individual path segments, Metropolis light transport \cite{veach1997metropolis} treats a complete path as a single sample in a global path space that is also used to define the target density. A new path is sampled from a Markov process by mutating an existing path according to a scalar contribution function. Various parameterizations of path space have been proposed to efficiently explore it using Markov chain methods~\cite{kelemen2002,jakob2012manifold,kaplanyan2014natural,li2015anisotropic}, and techniques have been proposed to combine different parameterizations to further improve the sampling process~\cite{Hachisuka14,otsu2017fusing,bitterli2017reversible}.
Our approach is similar to global path sampling techniques since we also importance sample entire paths. In particular, we sample paths in primary sample space as proposed by Kelemen et al.~\shortcite{kelemen2002}, which is simply a multi-dimensional unit hypercube. As in Kelemen's approach, we also build on existing path construction techniques that map primary sample space parameters into geometric paths, and that we treat as a black box. In contrast, our approach does not rely on Markov chain sampling, however. Instead, we obtain a non-linear, one-to-one mapping of primary sample space onto itself that produces the desired target density.

\subsection{Deep Learning to Sample Complex Data Distributions}

Recently, various neural network architectures have been proposed to learn generative models of high-dimensional data, such as images and videos. A generative model transforms random samples from a latent space into samples of some observable data, for example images, such that the distribution of the generated data matches the distribution of the observed data. Successful techniques include generative adversarial networks (GANs)~\shortcite{Goodfellow2014GAN} and variational autoencoders (VAEs)~\shortcite{Kingma2014VAE}.
Our idea is to use such a learned mapping from a latent space to a data space to perform importance sampling of light paths. A key requirement for this is that we need to be able to efficiently compute the Jacobian of this mapping, such that we can use it to perform a change of integration variables. However, the computation of the Jacobian involved in deep neural network is costly in techniques such as GANs or VAEs. Instead, we leverage a recent architecture called ``Real NVP''~\cite{DinhSB16} that was specifically designed as an invertible mapping whose Jacobian can be computed easily. A {concurrent work~\cite{Mueller18} also leverages ``Real NVP'' to construct an invertible mapping for use in Monte Carlo rendering. While they propose a slightly different representation of such a bijective mapping, in essence the concept is equivalent to ours.}

\subsection{Deep Learning in Monte Carlo Rendering}

Deep learning has been successfully applied to denoise Monte Carlo rendering~\cite{Bako:2017:KCN,Chaitanya17}, which is common in production today. Our approach is orthogonal to these techniques and can be combined with any denoiser. Neural networks have also been used as a regression architecture to interpolate radiance values from sparse samples~\cite{Ren:2013:GIR,Kallweit:2017:DSR}. These methods have no guarantees to converge to ground truth solutions, however, since they directly predict radiance based on a model learned from a limited amount of data. In contrast, we learn how to sample, and our sampling densities enable unbiased importance sampling. While we do not have a theoretical proof that our approach reduces variance, we demonstrate significant variance reduction in practical scenarios.

\section{Background} \label{sec:background}

The path integral formulation~\cite{veach1997metropolis} of the rendering equation~\cite{kajiya1986rendering} expresses the value $I_j$ of each pixel $j$ as an integral over the space of all light paths,
\begin{align}
I_{j}=\int_{\mathrm{\Omega}}f_j(x)\mathrm{d}\mu(x) \textrm{,}
\end{align}
where $x$ is a light path, $\mathrm{d}\mu$ is a measure on the space of paths, and $f_j$ is the measurement contribution function. In addition, $\Omega$ is the space of all light paths, consisting of the union of light paths of all lengths $k$, $\Omega = \cup_k \Omega_k$.

\begin{figure}
\centering
\includegraphics{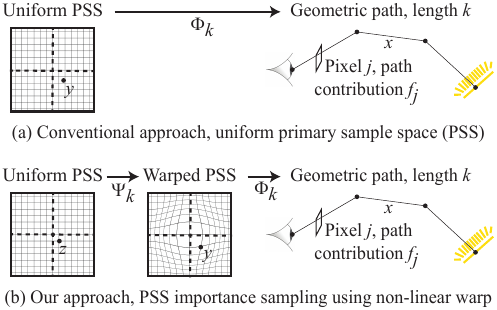}
\vspace{-1em}
\caption{\label{fig:overview}Comparison of the conventional approach and ours: (a) Usually, primary sample space (PSS) is sampled uniformly. Each point $y$ in primary sample space (PSS) of dimension $2(k+1)$ corresponds to a geometric path $x$ of length $k$ via a mapping $\Phi_k$. Conventional importance sampling is accounted for by the determinant of the Jacobian of $\Phi_k$. (b) Our approach introduces a non-linear warp $\Psi_k$ in primary sample space to further reduce variance, and we learn this mapping using a neural network.}
\vspace{-1em}
\end{figure}

An important observation is that there are many ways to parameterize light paths, and each parameterization has its own measure $\mathrm{d}\mu$. Considering only light paths of a certain length $k$, however, it is always possible to define a mapping $\Phi_k$ from a canonical parameterization over the $2(k+1)$-dimensional unit hypercube\footnote{This assumes that path vertices are on surfaces, and each of them is parameterized using two numbers. Depending on the details of the renderer, however, more parameters may be involved in practice, for example to select BRDF components.} to $\Omega_k$, that is, $\Phi_k :  [0,1]^{2(k+1)} \rightarrow \Omega_k$. This is also called primary sample space~\cite{kelemen2002}, and indeed each Monte Carlo rendering algorithm implicitly evaluates $\Phi_k$ when constructing light paths. Using $\Phi_k$ to perform a change of integration variables, we can rewrite the path integral formulation as
\begin{align}\label{eq:pssintegral}
I_{j,k} = \int_{\mathrm{\Omega}_k} f_{j}(x)\mathrm{d}\mu(x) = \int_{[0,1]^{2(k+1)}} f_{j}(\Phi_k(y)) \left| \frac{\partial \Phi_k(y)}{\partial y} \right|\mathrm{d}y,
\end{align}
where $I_{j,k}$ denotes the contribution of all paths of length $k$ to pixel $j$, $y \in [0,1]^{2(k+1)}$, and the notation $\left|\cdot\right|$ is shorthand for the determinant of the Jacobian matrix. Figure~\ref{fig:overview}(a) illustrates the notation in Equation~\ref{eq:pssintegral}.

In this view, incremental importance sampling techniques discussed in Section~\ref{sec:importancesampling} imply certain mappings $\Phi_k$ from primary sample space to geometric paths, and the effect of importance sampling is absorbed in the mapping $\Phi_k$. Usually, the integral in Equation~\ref{eq:pssintegral} is estimated using uniform sampling in primary sample space,
\begin{align}
\label{eq:pssestimate}
I_{j,k} \approx \frac{1}{N}\sum_{i=1}^N \frac{f_j(\Phi_k(y_i))}{\left| \frac{\partial \Phi_k(y_i)}{\partial y_i} \right|^{-1}},
\end{align}
where the $y_i \in [0,1]^{2(k+1)}$ are uniform random samples. The goal of importance sampling is to construct suitable mappings $\Phi_k$ with determinants that are inversely proportional to $f_j$ as much as possible\footnote{The determinant of the Jacobian is the inverse of the corresponding sampling density.}, which will reduce variance of the estimate in Equation~\ref{eq:pssestimate}. Incremental importance sampling approaches, however, rely only on local information about $f_j$, hence they are unable to account for non-local effects.

\section{Primary Sample Space (PSS) Warping} \label{sec:psswarping}

As illustrated in Figure~\ref{fig:overview}(b), the key idea of our approach is to introduce an additional mapping of primary sample space onto itself, acting as a non-uniform warp that leads to a non-uniform PSS density. We design this density to further reduce the variance of an existing renderer that evaluates Equation~\ref{eq:pssestimate}, which we treat as a black box. We learn the warp in a scene-dependent training phase, and then use it to draw as many samples for rendering as desired.

In the following, we describe how we construct the warp in the training phase. We start by introducing the details of our problem statement in Section~\ref{sec:problemformulation}, and explain in Section~\ref{sec:maximumlikelihood} how we achieve the objective of the PSS warp using an \emph{a posteriori} approach. This involves first drawing a number of samples from Equation~\ref{eq:pssestimate} and resampling them to obtain the desired target density. Then we use a maximum likelihood estimation of the warp to match the target density. In Section~\ref{sec:realnvp} we describe how we represent the warp in practice using the Real NVP architecture and solve the maximum likelihood estimation via gradient descent. We describe the details of the neural networks involved in Real NVP in Section~\ref{sec:stnetwork}, and describe how we use the warp during rendering in Section~\ref{sec:rendering}.

\subsection{Problem Formulation} \label{sec:problemformulation}

Let us denote the PSS warp that we will learn as
\begin{align}
\Psi_k : [0,1]^{2(k+1)} \rightarrow  [0,1]^{2(k+1)}.
\end{align}
Assuming this is bijective and differentiable, we can use it to introduce another change of integration variables  $y = \Psi_k(z)$ in Equation~\ref{eq:pssintegral},
\begin{align}
\label{eq:warpedpssintegral}
I_{j,k} = \int_{[0,1]^{2(k+1)}} f_{j}(\Phi_k(\Psi_k(z))) \left| \frac{\partial \Psi_k(z)}{\partial z} \right|\left| \frac{\partial \Phi_k(\Psi_k(z))}{\partial \Psi_k(z)} \right|\mathrm{d}z.
\end{align}
This leads to the Monte Carlo estimator
\begin{align}
\label{eq:warpedpssestimate}
I_{j,k} \approx \frac{1}{N} \sum_{i=1}^N \frac{f_j(\Phi_k(\Psi_k(z_i)))}
{\left| \frac{\partial \Psi_k(z_i)}{\partial z_i} \right|^{-1} \left| \frac{\partial \Phi_k(\Psi_k(z_i))}{\partial \Psi_k(z_i)} \right|^{-1}},
\end{align}
where the $z_i \in [0,1]^{2(k+1)}$ are uniform random samples. To reduce variance of this, we want to design $\Psi_k$ such that the integrand in Equation~\ref{eq:warpedpssintegral} is as close as possible to a constant. Note that the determinant of the Jacobian of our primary sample space warp, $\left| \partial \Psi_k(z)/\partial z \right|$ is the inverse of the desired non-uniform sampling density in a warped primary sample space, which serves as one part of the inputs to the black-box renderer.

\emph{Practical Simplifications.} Our description so far implies that we would need to learn a mapping $\Psi_k$ for each pixel. This is impractical, however, and we will firstly make the simplification to replace the measurement contribution function $f_j$ with the path throughput $f$, which is related to the measurement contribution by omitting the pixel filter (also called the importance function). While the path length $k$ is unlimited in theory, in practice, only paths with limited length are processed. {Additionally, longer paths generally contribute less to the final image than shorter paths, thus the actual advantages of importance sampling are reduced after the first few bounces. Hence, our second simplification is to only learn a PSS warp for certain limited path lengths $m$, and go on tracing paths with uniform random PSS coordinates for later bounces.}

\begin{figure}
\centering
\includegraphics{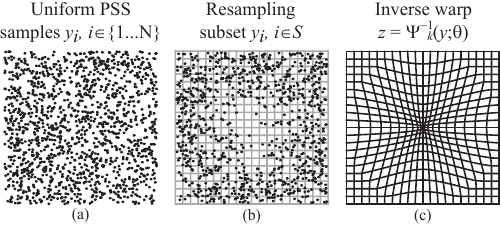}
\vspace{-1em}
\caption{We illustrate how we learn the inverse warp $\Psi^{-1}_k(y;\theta)$ to match a target density. (a) We first obtain uniform samples in PSS. (b) We resample them to obtain samples from the target density. (c) The inverse warp $\Psi^{-1}_k(y;\theta)$ is computed to maximize the likelihood of the target samples under the warp.}
\label{fig:resampling}
\end{figure}

\subsection{Maximum Likelihood Estimation of the PSS Warp}
\label{sec:maximumlikelihood}

As illustrated in Figure~\ref{fig:resampling}, we follow an \emph{a posteriori} approach to learn the PSS warp by first drawing a set of samples from a desired target density in primary sample space. Then we estimate the warp under a maximum likelihood objective for these observed samples.

\emph{Sampling the Target Density.} As mentioned above, in this paper we use the path throughput $f$ to define our target density as
\begin{align}
p_Y(y) = \frac{f(\Phi_k(y))}
{\left| \frac{\partial \Phi_k(y)}{\partial y} \right|^{-1}},
\end{align}
although any other target density could be used. {We obtain a set of samples $\mathbb{S}=\{y_{i}\:|\:i \in N^{+}, y_{i}\backsim p_Y\} $ from this distribution using a resampling process as proposed by Talbot et al.~\shortcite{Talbot:2005:IRG}. This involves first rendering a set $\mathbb{T}$ of candidate samples using a uniform PSS density as usual, and storing the PSS parameters and path throughputs for all samples. Then the subset $\mathbb{S}$ is extracted from $\mathbb{T}$ such that $\mathbb{S}$ follows the desired target density. The ratio of the sizes of $\mathbb{S}$ and $\mathbb{T}$ is determined by a parameter $\alpha$, that is, $|\mathbb{T}| = \alpha |\mathbb{S}|$}.

\emph{Maximum Likelihood Estimation.} We learn the inverse warp $z=\Psi_{k}^{-1}(y)$ using maximum likelihood estimation. Assume the warp is parameterized using parameters $\theta$, written as $z=\Psi_{k}^{-1}(y;\theta)$, $\theta\in\Theta$. Here, $\Theta$ represents the parameter space of a family of acceptable distributions. We first observe that the density $p_{Y}(\theta;y)$ and the warp $\Psi_{k}^{-1}$ are related by the change of variable formula
\begin{align}
    p_{Y}(\theta;y) = \left|\frac{\partial \Psi_{k}^{-1}(y;\theta)}{\partial y} \right|.
\end{align}
Note that this implies that latent variable $z=\Psi_k^{-1}(y)$ is distributed uniformly. Considering {$p_Y(\theta;y)$ a function of $\theta$, it is also called a likelihood function.} The maximum likelihood estimation approach tries to find the optimum parameters $\theta\in\Theta$ to maximize the likelihood, that the warp produces the data samples $y_i \backsim p_Y $ from uniform samples $z_i$. More precisely, maximizing the log-likelihood of the data samples { $y_i, i \in \{1,\dots,|\mathbb{S}|\}$} can be written as
\begin{align}
\label{eq:ML}
\theta^*
= \argmax_{\theta\in\Theta} \sum_{i} \log\left( \left|\frac{\partial \Psi^{-1}(y_i;\theta)}{\partial y_i} \right|\right).
\end{align}

\begin{figure}
\centering
\includegraphics[width=\columnwidth]{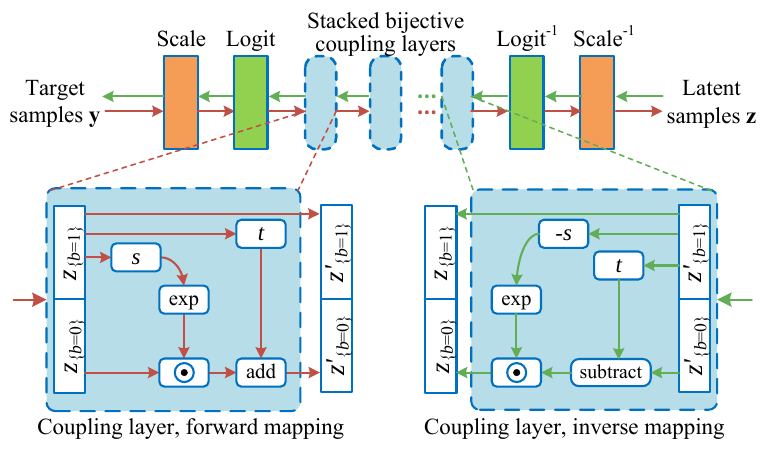}
\vspace{-1em}
\caption{Computational structure of our neural importance sampling model based on Real NVP. The core element of this architecture consists of a set of stacked, invertible coupling layers. The functions $s$ and $t$ are implemented using neural networks as shown in Figure~\ref{fig:stnet}, and stacking several coupling layers makes it possible to represent complex, bijective mappings.}
\label{fig:realNVP}
\end{figure}

\subsection{PSS Warping using Real NVP} \label{sec:realnvp}

We are leveraging a recent neural network architecture dubbed as ``Real NVP'' transformation~\cite{DinhSB16} to represent our PSS warp $\Psi$ and its inverse $\Psi^{-1}$. Real NVP provides desired properties that fit into our problem: it is guaranteed to be bijective, it is easy to invert, the determinant of its Jacobian can be evaluated efficiently, and maximum likelihood estimation can be performed via gradient descent.

\begin{figure}
\centering
\includegraphics[width=0.60\linewidth]{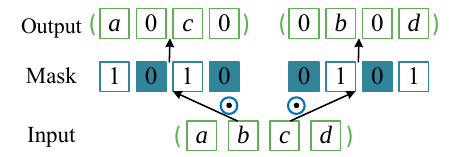}
\caption{\label{fig:mask}Element-wise masking of a 4D input vector. Masks are designed as a binary ``checkerboard'' mode, {where elements of the mask are set to alternating 0 and 1.}}
\vspace{-1em}
\end{figure}

In a nutshell, Real NVP transformations consist of multiple stacked (concatenated) so-called affine coupling layers as shown in Figure~\ref{fig:realNVP}, top. The mapping computed by each coupling layer is designed to have all the properties mentioned above, and the concatenation of multiple coupling layers can represent complex mappings. Assume each coupling layer computes a mapping from a $D$-dimensional space onto itself. A key idea is to split the input vector $z$ into two disjoint parts, which we represent by a binary mask $b$ of size $D$. {An example of applying element-wise masking to a 4D vector} is shown in Figure~\ref{fig:mask}. Let $\{b=1\}$ be the set of indices where the mask has value $1$, and similar for $\{b=0\}$.
Each coupling layer forwards the first part of the input $z_{\{b=1\}}$ directly to its output. In addition, the second part of the output $z_{\{b=0\}}$ consists of an affine mapping that is constructed using functions $s(z_{\{b=1\}})$ and $t(z_{\{b=1\}})$ of the first part of the input. As shown at the bottom left of Figure~\ref{fig:realNVP}, a coupling layer computes
\begin{align} \label{eq:biject}
z'_{\{b=1\}} &= z_{\{b=1\}} \nonumber \\
z'_{\{b=0\}} &= z_{\{b=0\}} \odot \exp (s(z_{\{b=1\}}))+ t(z_{\{b=1\}}),
\end{align}
where $s$ and $t$ are functions from $\mathbb{R}^{|\{b=1\}|} \rightarrow \mathbb{R}^{|\{b=0\}|}$, and $\odot$ is the element-wise product. Crucially, such a coupling layer is trivial to invert as shown at the bottom right of Figure~\ref{fig:realNVP},
\begin{align} \label{eq:invbiject}
z_{\{b=1\}} &= z'_{\{b=1\}} \nonumber \\
z_{\{b=0\}} &= \left(z'_{\{b=0\}}-t(z_{\{b=1\}})\right) \odot \exp (-s(z_{\{b=1\}})) ,
\end{align}
and its Jacobian is a triangular matrix whose determinant is also trivial to obtain as $\exp\left(\sum_j s(z_{\{b=1\}})_j \right)$~\cite{DinhSB16}. Neither operation requires inverting the Jacobians of $s$ and $t$ nor computing their inverses. Hence they can be arbitrarily complex, and we implement them using deep neural networks. We describe our network architecture to implement $s$ and $t$ in more detail below. Finally, we apply a linear scaling and a logit mapping at the start, and their inverse at the end of the model. {Linear scaling layers are employed to ensure that values are within valid ranges.} We found that this improves training convergence of our model. 

\begin{figure}
\centering
    \includegraphics[width=1.0\linewidth]{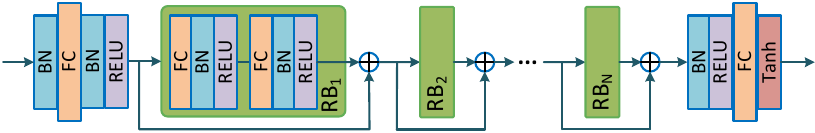}
\caption{The neural network architecture to compute the $s$ and $t$ functions in a bijective coupling layer consists of $N$ residual blocks (RB), containing fully connected layers (FC), batch normalization layers (BN), and ReLU activation units. Additionally, we have fully connected, batch normalization, and activation layers before and after the residual blocks. All fully connected layers contain the same number of neurons.}
\label{fig:stnet}
\vspace{-2em}
\end{figure}

In summary, we implement our PSS warp $\Psi$ and its inverse using Real NVP with multiple stacked affine coupling layers. Since each bijective coupling layer only warps part of the input dimensions, we concatenate several coupling layers to warp all dimensions non-linearly. In our experiments, we use eight coupling layers and masks where either the even or odd bits are set to zero or one, respectively. We also experimented with other masks, but did not observe any significant differences. The warp parameters $\theta$ in Equation~\ref{eq:ML} correspond to the trainable weights of the neural networks that define the $s$ and $t$ functions in all coupling layers (each layer has its own $s$ and $t$ functions, that is, neural networks). We describe our networks in more detail in the next section. We perform maximum likelihood estimation of $\Psi^{-1}$ using gradient descent and standard backpropagation techniques for neural networks.

\subsection{Neural Network Architecture for Coupling Layers} \label{sec:stnetwork}

Figure~\ref{fig:stnet} shows the neural network architecture that we use to compute both the $s$ and $t$ functions in Equations~\ref{eq:biject} and \ref{eq:invbiject}. Each function in each coupling layer has its own set of trainable network weights. The network consists of several blocks with residual connections~\cite{He16}. Each block contains two fully connected layers (FC) with batch normalization (BN) and rectified linear unit (ReLU) activation $\sigma(x)=max(0, x)$, and a block can be bypassed via a residual connection. In addition, there are more such layers before and after the residual blocks. The warping capability of the model is directly affected by the number of coupling layers. More coupling layers enable the model to learn complex mappings. Meanwhile, using more coupling layers and residual blocks will lead to relatively long training time. We discuss the exact network parameters in Section~\ref{sec:results}.

\subsection{Generating Rendering Samples} \label{sec:rendering}

\begin{figure}
\centering
\includegraphics{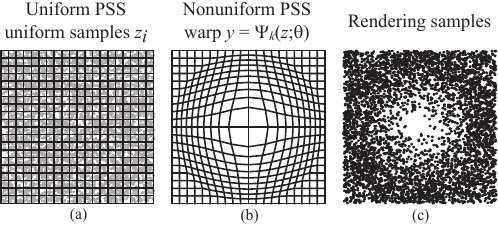}
\vspace{-1em}
\caption{\label{fig:samplegeneration}During rendering, we use the forward warp $\Psi$ to draw an arbitrary number of samples distributed approximately according to the target density as shown in this 2D illustration. (a) Uniform samples in PSS. (b) Warped PSS using $\Psi$. (c) Non-uniform samples in PSS, which we provide as input to a black-box renderer.}
\vspace{-1em}
\end{figure}

For rendering, we use the forward mapping $\Psi$, which can easily be derived from $\Psi^{-1}$ as shown above. Note that $\Psi$ reuses trained parameters of $\Psi^{-1}$. We generate an arbitrary number of samples distributed approximately according to the target density as represented by the learned warp, as illustrated in Figure~\ref{fig:samplegeneration}.

{In this paper, path tracing (PT) with next event estimation is utilized as the underlying rendering algorithm to show how it can be improved via our method}. The construction of paths will consume PSS coordinates. Based on the practical simplifications in Section~\ref{sec:problemformulation}, we opt to learn to importance sample dimensions of the first $m$ bounces (e.g. $m = 1,2,3,\dots$). After $m$ bounces, we continue to trace the path using uniform random PSS coordinates.

\section{{Implementation Details}} \label{sec:implementation}

In this section, we describe datasets used for training, validation and test. We also provide details of neural network training.

\textbf{Training Datasets.} \,We set the number of examples (that is, samples $y_i$ in Equation~\ref{eq:ML} in a training data set) to $k$ times a certain image resolution, equivalent to $k$ examples per pixel (\emph{epp}) on average. {For brevity, we will denote a dataset as \textit{epp}-k. When generating training data we use small image resolutions. As shown in Table~\ref{tab:trainingtime}, we chose $160 \times 88$ for the first four scenes, and $100 \times 100$ for the \textsc{Torus} scene. Note that final images can be rendered at any desired higher resolution. We empirically found that good resampling results are obtatined when $\alpha$ is at least 6, and all experiments follow this heuristic.} 
During training, $80\%$ examples of a training dataset are randomly selected as training data and the rest are used for validation. Simultaneously, we provide an accompanying test dataset to test the performance of the neural networks. None of the test data are used in the training and validation process.

\textbf{Model Initialization.} \,For faster convergence during scene-dependent training, we pre-train our neural network to achieve an identity warp. We use the resulting network weights as initialization for scene-dependent training, instead of the usual random or Xavier initialization~\shortcite{glorot10}. We have observed that this leads to faster scene-dependent training convergence in practice.

\textbf{Training.} \,We train neural networks corresponding to inverse warp $\Psi^{-1}$ in an end-to-end fashion using the TensorFlow \cite{tf15} framework. The networks are optimized using Adam~\shortcite{Kingma14} with a learning rate $10^{-4}$ and decay rates $\beta_{1}=0.9, \beta_{2}=0.99$. Training examples are fed into the neural network in mini-batches size of 2000, and the order of examples in a mini-batch is randomly shuffled in each iteration.

\section{{Results and Analysis}} \label{sec:results}

All our experiments are conducted on a workstation with an octa-core 3.60 GHz i7-7700 CPU and Nvidia GeForce GTX 1070 GPU. We implement our method based on PBRT~\cite{pbrt}, and execute it in a hybrid way using both GPU and CPU. We perform neural network training and evaluation on the GPU and the renderer runs on the CPU. {We firstly validate the neural network designs in Section~\ref{sec:validatearch} and \ref{sec:datasize}, where experiments are in terms of 4D PSS learning for the \textsc{Country Kitchen} scene. Then we apply our approach to five scenes with a range of challenging settings.}

\subsection{Neural Network Architecture Validation} \label{sec:validatearch}

\begin{figure}
\begin{center}
\begin{tabular}{@{}c@{}c@{}}
\begin{minipage}[t]{0.5\linewidth}
    \centering
    \includegraphics[width=1.0\linewidth]{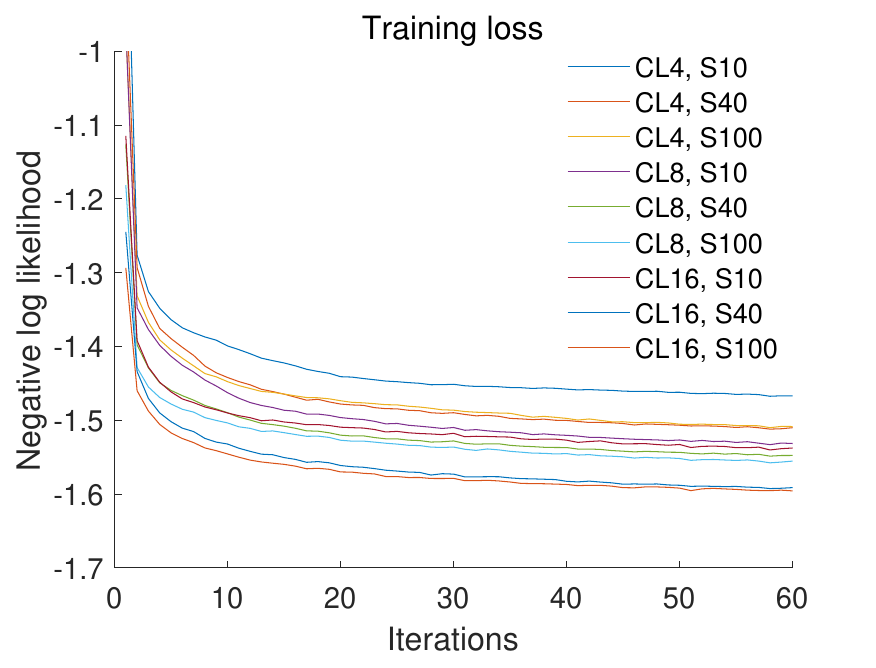}
\end{minipage}&
\begin{minipage}[t]{0.5\linewidth}
    \centering
    \includegraphics[width=1.0\linewidth]{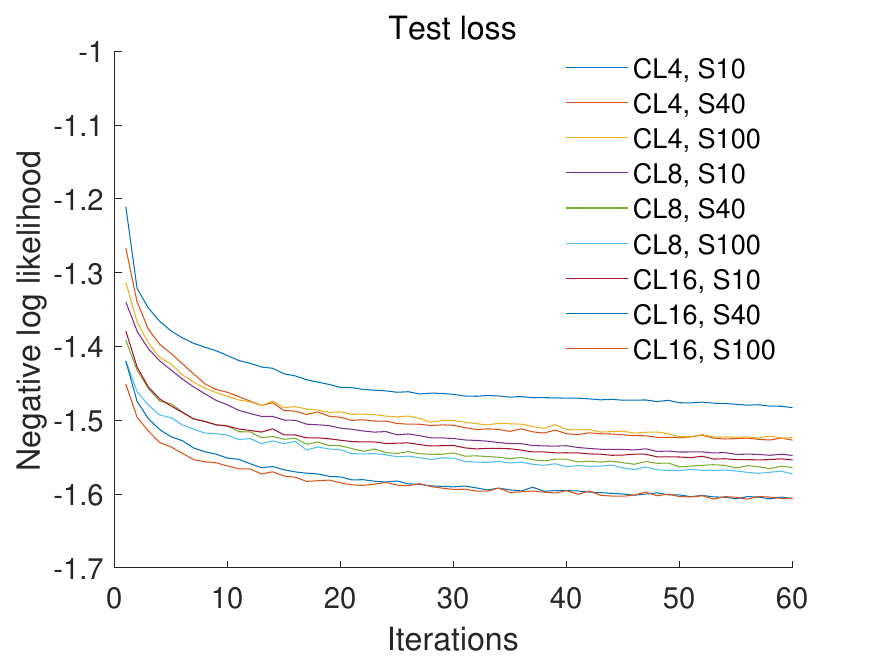}
\end{minipage}\\
\end{tabular}
\end{center}
\vspace{-0.5em}
\caption{\label{fig:validate_arch}Training and test loss of different neural network architectures. CL-$i$, S-$j$ denotes $i$ coupling layers and $j$ neurons per hidden layer. The residual block count is set to 2 {and training datasets use \textit{epp}-16.}}
\vspace{-1.5em}
\end{figure}

The capacity of our neural network to represent complex mappings is related to the number of coupling layers and the size of a hidden layer. To get an efficient sampling model, we tend to use a lightweight architecture with conservative numbers of coupling layers. In Figure~\ref{fig:validate_arch}, we compare different combinations of coupling layer count (CL) and hidden layer size (S). Each training session runs for 60 epochs. As can be seen, more coupling layers lead to lower loss values. Given a fixed number of coupling layers, increasing the number of neurons of hidden layers further reduce loss values. The deep and wide architecture provides the lowest loss values, while it incurs a long training time of several hours. We empirically find that eight coupling layers and 40 neurons per hidden layer can successfully learn 4D to 8D target densities in our experiments and keep sampling cost small.

\subsection{Training Data Size and Density Error} \label{sec:datasize}

Training data provides an implicit description of the target probability density function. To examine the effects of different numbers of training examples, we compare training and test losses with respect to training datasets of five different sizes. During model training, we use a low image resolution of $160 \times 88$ pixels. The dataset with the smallest size has one example for each pixel on average. 

\begin{figure}
\begin{center}
\begin{tabular}{@{}c@{}c@{}}
\begin{minipage}[c]{0.50\linewidth}
    \centering
    \includegraphics[width=1.0\linewidth]{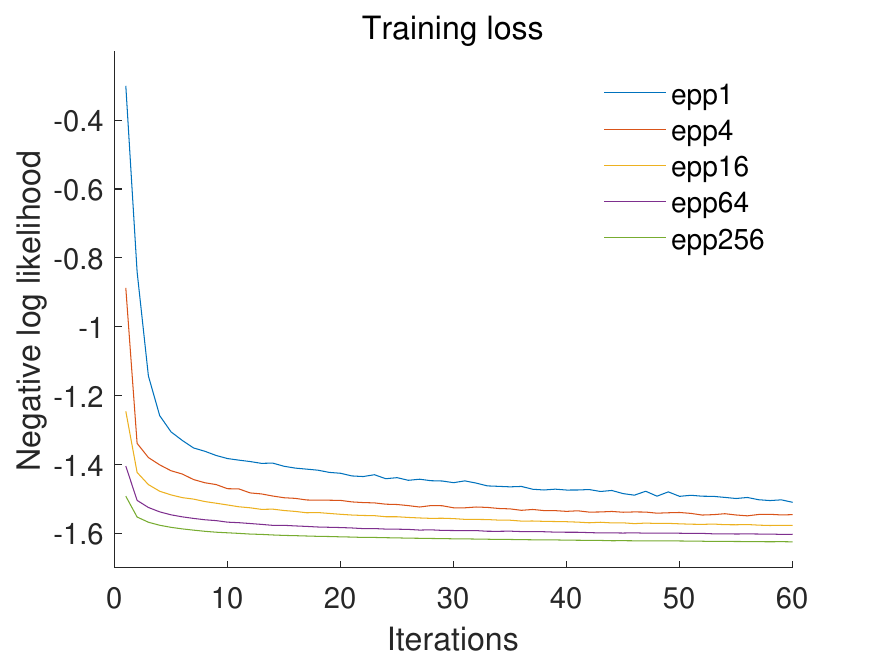}
\end{minipage}&
\begin{minipage}[c]{0.50\linewidth}
    \centering
    \includegraphics[width=1.0\linewidth]{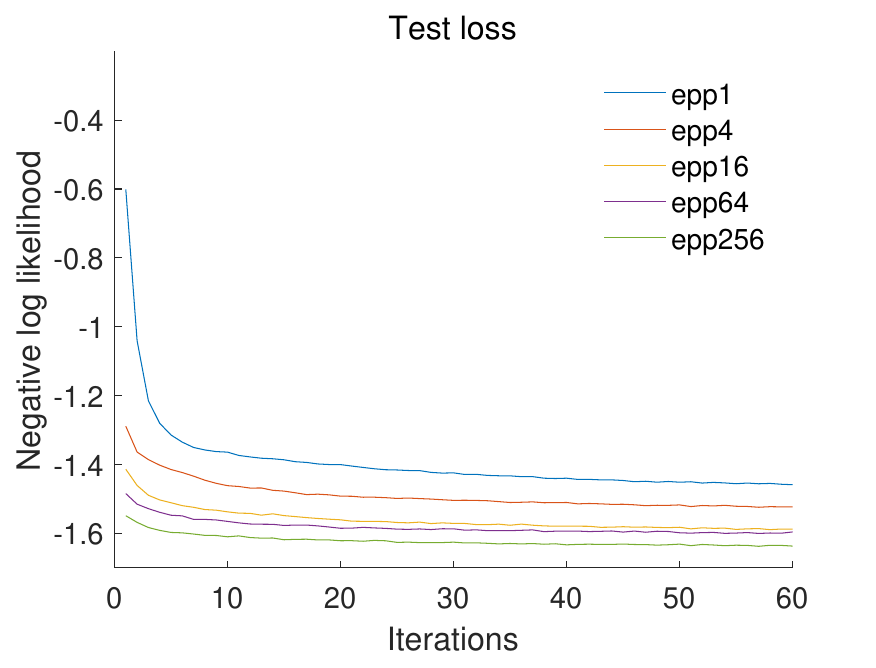}
\end{minipage}\\
\end{tabular}
\end{center}
\vspace{-0.5em}
\caption{\label{fig:validate_datasize}Comparisons of training loss and test loss with respect to the size of training datasets. Sizes ranges from \textit{epp}-1 to \textit{epp}-256.}
\vspace{-1em}
\end{figure}

Figure~\ref{fig:validate_datasize} plots the training and test losses for each dataset. As shown, a lower negative log likelihood can be obtained with an increasing size of datasets, and the improvements brought by using more training data gradually reduce beyond \emph{epp} 16.

\begin{figure}
\begin{center}
\begin{tabular}{@{}c@{}c@{}}
\begin{minipage}[t]{0.50\linewidth}
    \centering
    \includegraphics[width=1.0\linewidth]{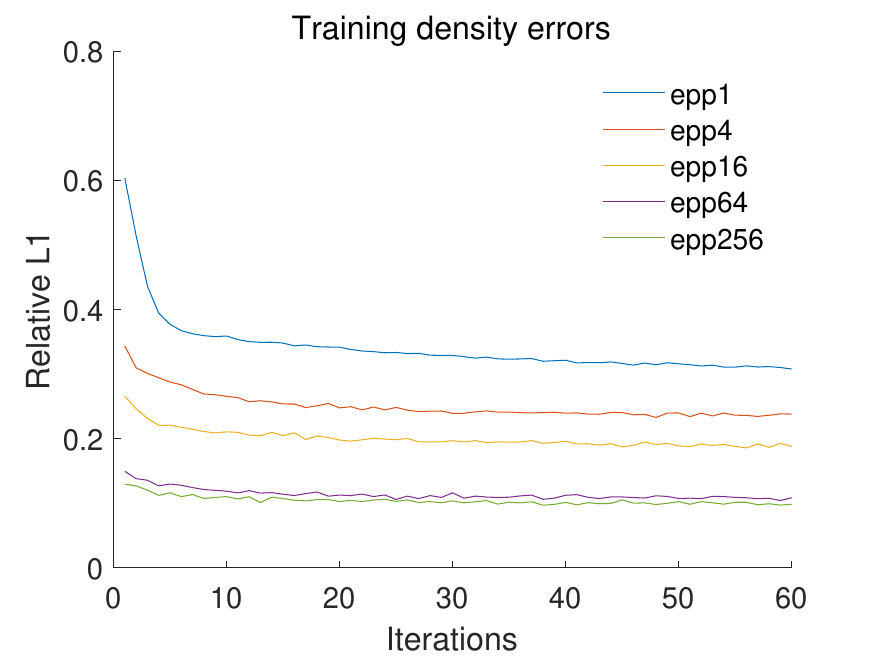}
\end{minipage}&
\begin{minipage}[t]{0.50\linewidth}
    \centering
    \includegraphics[width=1.0\linewidth]{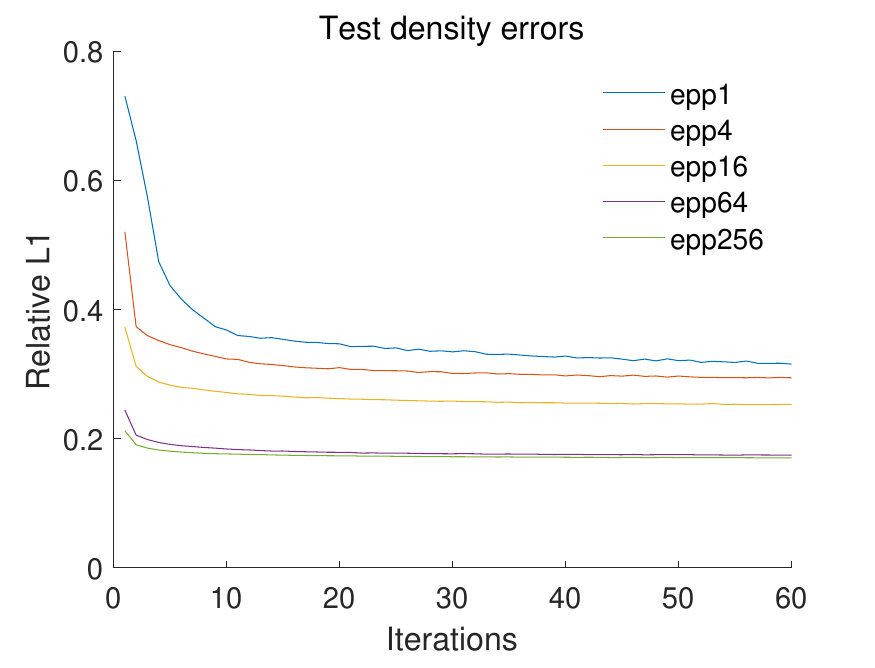}
\end{minipage}\\
\end{tabular}
\end{center}
\vspace{-0.5em}
\caption{Relative $L_1$ density error for training and test processes, with respect to different sizes of data sets. Each training session runs for 60 epochs.}
\label{fig:density_relL1}
\vspace{-2em}
\end{figure}

In addition, Figure~\ref{fig:density_relL1} shows relative $L_1$ errors of density values for both training and test dataset. The relative density errors is computed with
\begin{equation}
\epsilon=\frac{1}{N}\sum_{i=1}^{N}\frac{\lvert P(D_{i})-T_{i}\rvert}{max(P(D_{i}), T_{i})} \textrm{.}
\end{equation}
Here, $P(D_{i})$ stands for the density deduced by neural networks, $T_{i}$ represents the target density value of $D_{i}$ and  $\left|\cdot\right|$ indicates the absolute value. Target densities are readily obtained up to a normalizing constant in the resampling process. Using more training data helps to reduce relative density errors, but the improvement of density accuracy gets smaller with increasing size of datasets.

\subsection{Rendering Results and Further Analysis}

\begin{figure*}[p]
    \makebox[\h][c]{}\hfill
    \raisebox{1.0mm}[0pt][0pt]{
    \makebox[\hh][c]{\small{PT no warp}}\hfill
    \mbox{\hspace*{0.0mm}}
    \makebox[\hh][c]{\small{Kdtree warp}}\hfill
    \mbox{\hspace*{0.0mm}}
    \makebox[\hh][c]{\small{Our 4D}}\hfill
    \mbox{\hspace*{0.0mm}}
    \makebox[\hh][c]{\small{Our 6D}}\hfill
    \mbox{\hspace*{0.0mm}}
    \makebox[\hh][c]{\small{Our 8D}}\hfill
    \mbox{\hspace*{0.0mm}}
    \makebox[\hh][c]{\small{Reference}}\hfill
    }\\
    \includegraphics[width=\h]{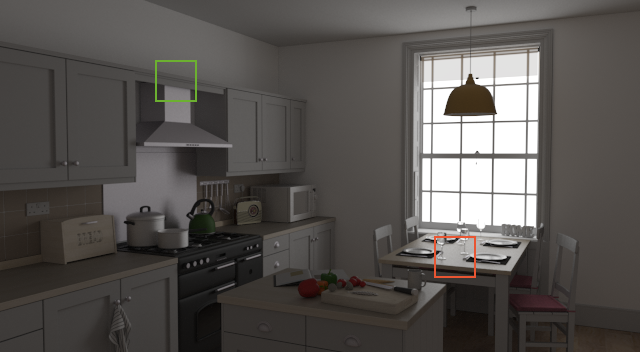}\\
    \makebox[\h][r]{\textsc{Country Kitchen}\hspace{10mm}\footnotesize{MSE}} \hfill
    \raisebox{21.75mm}[0pt][0pt]{
    \includegraphics[width=\hh]{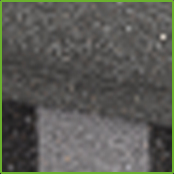}\hfill
    \mbox{\hspace*{0.0mm}}
    \includegraphics[ width=\hh]{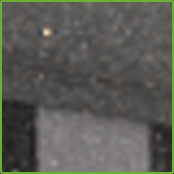}\hfill
    \mbox{\hspace*{0.0mm}}
    \includegraphics[ width=\hh]{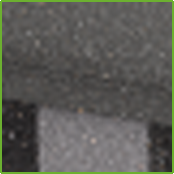}\hfill
    \mbox{\hspace*{0.0mm}}
    \includegraphics[ width=\hh]{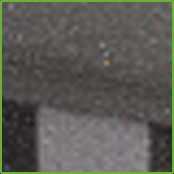}\hfill
    \mbox{\hspace*{0.0mm}}
    \includegraphics[ width=\hh]{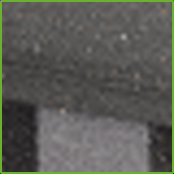}\hfill
    \mbox{\hspace*{0.0mm}}
    \includegraphics[ width=\hh]{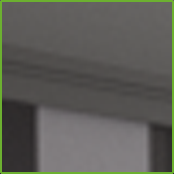}\hfill
    }\\
    \makebox[\h][r]{\footnotesize{1-SSIM}} \hfill
    \raisebox{7.75mm}[0pt][0pt]{
    \includegraphics[ width=\hh]{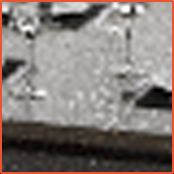}\hfill
    \mbox{\hspace*{0.0mm}}
    \includegraphics[ width=\hh]{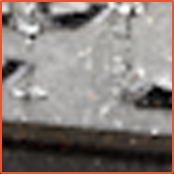}\hfill
    \mbox{\hspace*{0.0mm}}
    \includegraphics[ width=\hh]{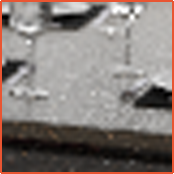}\hfill
    \mbox{\hspace*{0.0mm}}
    \includegraphics[ width=\hh]{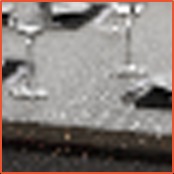}\hfill
    \mbox{\hspace*{0.0mm}}
    \includegraphics[ width=\hh]{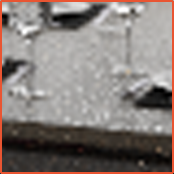}\hfill
    \mbox{\hspace*{0.0mm}}
    \includegraphics[ width=\hh]{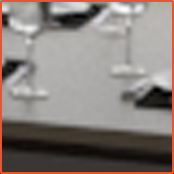}\hfill
    }\\
    \makebox[\h][c]{} \hfill
    \raisebox{8.0mm}[0pt][0pt]{
    \makebox[\hh][c]{\footnotesize{0.0401}}\hfill
    \mbox{\hspace*{0.0mm}}
    \makebox[\hh][c]{\footnotesize{0.0220}}\hfill
    \mbox{\hspace*{0.0mm}}
    \makebox[\hh][c]{\footnotesize{0.0170}}\hfill
    \mbox{\hspace*{0.0mm}}
    \makebox[\hh][c]{\footnotesize{0.0172}}\hfill
    \mbox{\hspace*{0.0mm}}
    \makebox[\hh][c]{\footnotesize{0.0205}}\hfill
    \mbox{\hspace*{0.0mm}}
    \makebox[\hh][c]{\footnotesize{}}\hfill
    }\\
    \makebox[\h][c]{} \hfill
    \raisebox{8.0mm}[0pt][0pt]{
    \makebox[\hh][c]{\footnotesize{0.4097}}\hfill
    \mbox{\hspace*{0.0mm}}
    \makebox[\hh][c]{\footnotesize{0.3861}}\hfill
    \mbox{\hspace*{0.0mm}}
    \makebox[\hh][c]{\footnotesize{0.3591}}\hfill
    \mbox{\hspace*{0.0mm}}
    \makebox[\hh][c]{\footnotesize{0.3571}}\hfill
    \mbox{\hspace*{0.0mm}}
    \makebox[\hh][c]{\footnotesize{0.3538}}\hfill
    \mbox{\hspace*{0.0mm}}
    \makebox[\hh][c]{\footnotesize{}}\hfill
    }\\
    \vspace{-11.5mm}\\
    \includegraphics[width=\h]{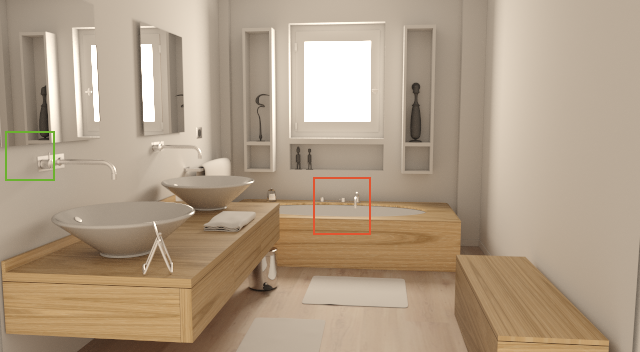}\\
    \makebox[\h][r]{\textsc{Salle De Bain}\hspace{13mm}\footnotesize{MSE}} \hfill
    \raisebox{21.75mm}[0pt][0pt]{
    \includegraphics[ width=\hh]{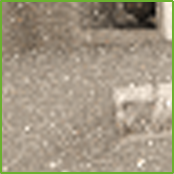}\hfill       \mbox{\hspace*{0.0mm}}
    \includegraphics[ width=\hh]{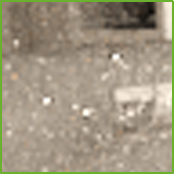}\hfill       \mbox{\hspace*{0.0mm}}
    \includegraphics[ width=\hh]{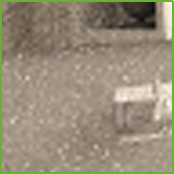}\hfill       \mbox{\hspace*{0.0mm}}
    \includegraphics[ width=\hh]{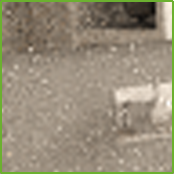}\hfill       \mbox{\hspace*{0.0mm}}
    \includegraphics[ width=\hh]{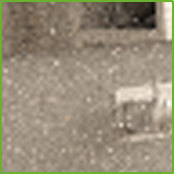}\hfill
    \mbox{\hspace*{0.0mm}}
    \includegraphics[ width=\hh]{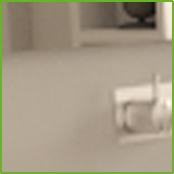}\hfill
    }\\
    \makebox[\h][r]{\footnotesize{1-SSIM}} \hfill
    \raisebox{7.75mm}[0pt][0pt]{
    \includegraphics[ width=\hh]{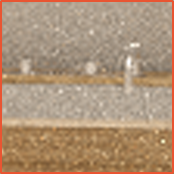}\hfill       \mbox{\hspace*{0.0mm}}
    \includegraphics[ width=\hh]{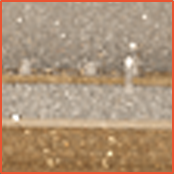}\hfill       \mbox{\hspace*{0.0mm}}
    \includegraphics[ width=\hh]{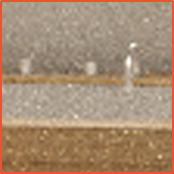}\hfill       \mbox{\hspace*{0.0mm}}
    \includegraphics[ width=\hh]{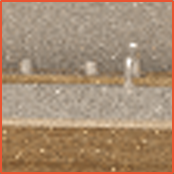}\hfill       \mbox{\hspace*{0.0mm}}
    \includegraphics[ width=\hh]{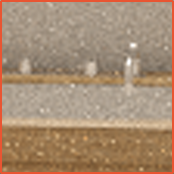}\hfill       \mbox{\hspace*{0.0mm}}
    \includegraphics[ width=\hh]{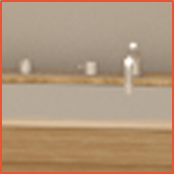}\hfill
    }\\
    \makebox[\h][c]{} \hfill
    \raisebox{8mm}[0pt][0pt]{
    \makebox[\hh][c]{\footnotesize{0.2080}}\hfill
    \mbox{\hspace*{0.0mm}}
    \makebox[\hh][c]{\footnotesize{0.2687}}\hfill
    \mbox{\hspace*{0.0mm}}
    \makebox[\hh][c]{\footnotesize{0.0859}}\hfill
    \mbox{\hspace*{0.0mm}}
    \makebox[\hh][c]{\footnotesize{0.0813}}\hfill
    \mbox{\hspace*{0.0mm}}
    \makebox[\hh][c]{\footnotesize{0.1178}}\hfill
    \mbox{\hspace*{0.0mm}}
    \makebox[\hh][c]{\footnotesize{}}\hfill
    }\\
    \makebox[\h][c]{} \hfill
    \raisebox{8mm}[0pt][0pt]{
    \makebox[\hh][c]{\footnotesize{0.4561}}\hfill
    \mbox{\hspace*{0.0mm}}
    \makebox[\hh][c]{\footnotesize{0.4919}}\hfill
    \mbox{\hspace*{0.0mm}}
    \makebox[\hh][c]{\footnotesize{0.4000}}\hfill
    \mbox{\hspace*{0.0mm}}
    \makebox[\hh][c]{\footnotesize{0.4034}}\hfill
    \mbox{\hspace*{0.0mm}}
    \makebox[\hh][c]{\footnotesize{0.4132}}\hfill
    \mbox{\hspace*{0.0mm}}
    \makebox[\hh][c]{\footnotesize{}}\hfill
    }\\
    \vspace{-11.5mm}\\
    \includegraphics[width=\h]{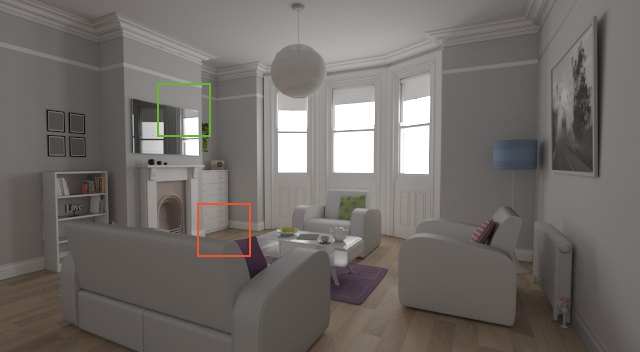}\\
    \makebox[\h][r]{\textsc{White Room}\hspace{14mm}\footnotesize{MSE}} \hfill
    \raisebox{21.75mm}[0pt][0pt]{
    \includegraphics[ width=\hh]{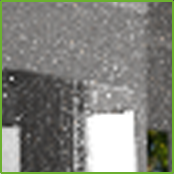}\hfill
    \mbox{\hspace*{0.0mm}}
    \includegraphics[ width=\hh]{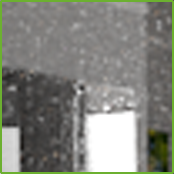}\hfill       \mbox{\hspace*{0.0mm}}
    \includegraphics[ width=\hh]{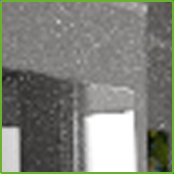}\hfill       \mbox{\hspace*{0.0mm}}
    \includegraphics[ width=\hh]{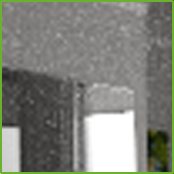}\hfill       \mbox{\hspace*{0.0mm}}
    \includegraphics[ width=\hh]{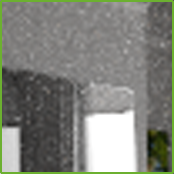}\hfill       \mbox{\hspace*{0.0mm}}
    \includegraphics[ width=\hh]{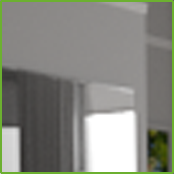}\hfill
    }\\
    \makebox[\h][r]{\footnotesize{1-SSIM}} \hfill
    \raisebox{7.75mm}[0pt][0pt]{
    \includegraphics[ width=\hh]{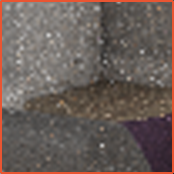}\hfill       \mbox{\hspace*{0.0mm}}
    \includegraphics[ width=\hh]{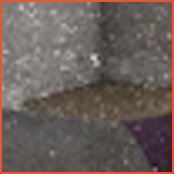}\hfill       \mbox{\hspace*{0.0mm}}
    \includegraphics[ width=\hh]{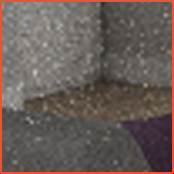}\hfill       \mbox{\hspace*{0.0mm}}
    \includegraphics[ width=\hh]{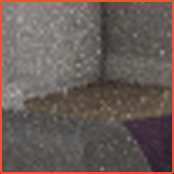}\hfill       \mbox{\hspace*{0.0mm}}
    \includegraphics[ width=\hh]{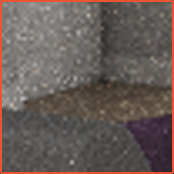}\hfill       \mbox{\hspace*{0.0mm}}
    \includegraphics[ width=\hh]{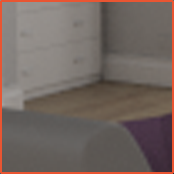}\hfill
    }\\
    \makebox[\h][c]{} \hfill
    \raisebox{8mm}[0pt][0pt]{
    \makebox[\hh][c]{\footnotesize{0.2155}}\hfill
    \mbox{\hspace*{0.0mm}}
    \makebox[\hh][c]{\footnotesize{0.1235}}\hfill
    \mbox{\hspace*{0.0mm}}
    \makebox[\hh][c]{\footnotesize{0.0820}}\hfill
    \mbox{\hspace*{0.0mm}}
    \makebox[\hh][c]{\footnotesize{0.0677}}\hfill
    \mbox{\hspace*{0.0mm}}
    \makebox[\hh][c]{\footnotesize{0.0832}}\hfill
    \mbox{\hspace*{0.0mm}}
    \makebox[\hh][c]{\footnotesize{}}\hfill
    }\\
    \makebox[\h][c]{} \hfill
    \raisebox{8mm}[0pt][0pt]{
    \makebox[\hh][c]{\footnotesize{0.4876}}\hfill
    \mbox{\hspace*{0.0mm}}
    \makebox[\hh][c]{\footnotesize{0.4308}}\hfill
    \mbox{\hspace*{0.0mm}}
    \makebox[\hh][c]{\footnotesize{0.4231}}\hfill
    \mbox{\hspace*{0.0mm}}
    \makebox[\hh][c]{\footnotesize{0.3773}}\hfill
    \mbox{\hspace*{0.0mm}}
    \makebox[\hh][c]{\footnotesize{0.4275}}\hfill
    \mbox{\hspace*{0.0mm}}
    \makebox[\hh][c]{\footnotesize{}}\hfill
    }\\
    \vspace{-11.5mm}\\
    \includegraphics[width=\h]{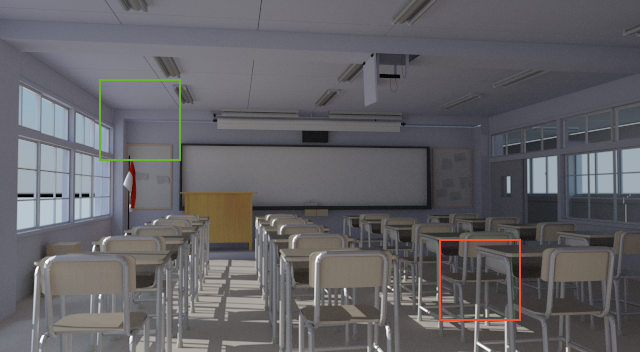}\\
    \makebox[\h][r]{\textsc{Classroom}\hspace{15mm}\footnotesize{MSE}} \hfill
    \raisebox{21.75mm}[0pt][0pt]{
    \includegraphics[ width=\hh]{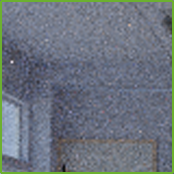}\hfill
    \mbox{\hspace*{0.0mm}}
    \includegraphics[ width=\hh]{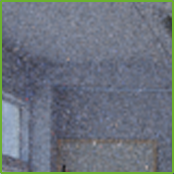}\hfill
    \mbox{\hspace*{0.0mm}}
    \includegraphics[ width=\hh]{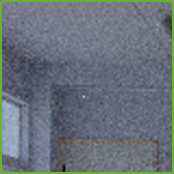}\hfill
    \mbox{\hspace*{0.0mm}}
    \includegraphics[ width=\hh]{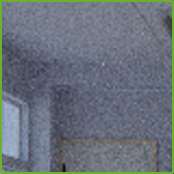}\hfill
    \mbox{\hspace*{0.0mm}}
    \includegraphics[ width=\hh]{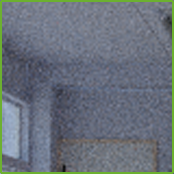}\hfill
    \mbox{\hspace*{0.0mm}}
    \includegraphics[ width=\hh]{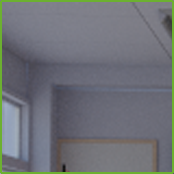}\hfill
    }\\
    \makebox[\h][r]{\footnotesize{1-SSIM}} \hfill
    \raisebox{7.75mm}[0pt][0pt]{
    \includegraphics[ width=\hh]{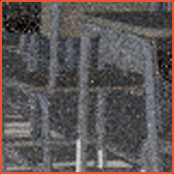}\hfill
    \mbox{\hspace*{0.0mm}}
    \includegraphics[ width=\hh]{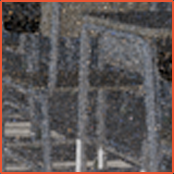}\hfill
    \mbox{\hspace*{0.0mm}}
    \includegraphics[ width=\hh]{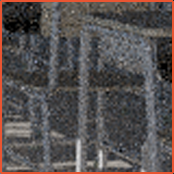}\hfill
    \mbox{\hspace*{0.0mm}}
    \includegraphics[ width=\hh]{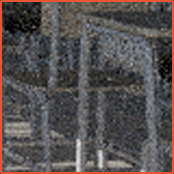}\hfill
    \mbox{\hspace*{0.0mm}}
    \includegraphics[ width=\hh]{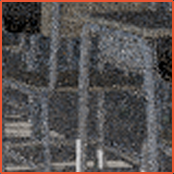}\hfill
    \mbox{\hspace*{0.0mm}}
    \includegraphics[ width=\hh]{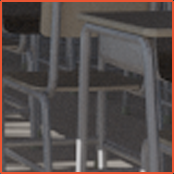}\hfill
    }\\
    \makebox[\h][c]{} \hfill
    \raisebox{8mm}[0pt][0pt]{
    \makebox[\hh][c]{\footnotesize{0.01277}}\hfill
    \mbox{\hspace*{0.0mm}}
    \makebox[\hh][c]{\footnotesize{0.01295}}\hfill
    \mbox{\hspace*{0.0mm}}
    \makebox[\hh][c]{\footnotesize{0.01314}}\hfill
    \mbox{\hspace*{0.0mm}}
    \makebox[\hh][c]{\footnotesize{0.01329}}\hfill
    \mbox{\hspace*{0.0mm}}
    \makebox[\hh][c]{\footnotesize{0.01326}}\hfill
    \mbox{\hspace*{0.0mm}}
    \makebox[\hh][c]{\footnotesize{}}\hfill
    }\\
    \makebox[\h][c]{} \hfill
    \raisebox{8mm}[0pt][0pt]{
    \makebox[\hh][c]{\footnotesize{0.4639}}\hfill
    \mbox{\hspace*{0.0mm}}
    \makebox[\hh][c]{\footnotesize{0.4387}}\hfill
    \mbox{\hspace*{0.0mm}}
    \makebox[\hh][c]{\footnotesize{0.4876}}\hfill
    \mbox{\hspace*{0.0mm}}
    \makebox[\hh][c]{\footnotesize{0.4905}}\hfill
    \mbox{\hspace*{0.0mm}}
    \makebox[\hh][c]{\footnotesize{0.4877}}\hfill
    \mbox{\hspace*{0.0mm}}
    \makebox[\hh][c]{\footnotesize{}}\hfill
    }\\
    \vspace{-11.5mm}\\
    \adjustbox{bgcolor={rgb}{0.3 0.3 0.3}}{\includegraphics[trim={-177 0 -177 0}, clip, width=\h]{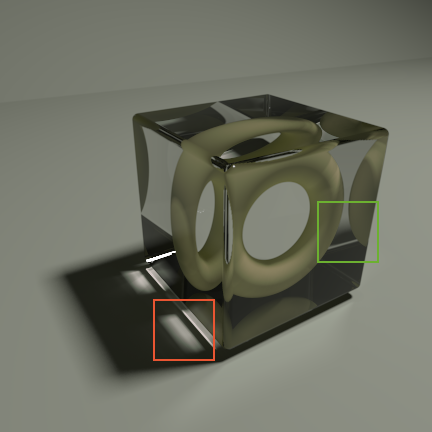}}\\
    \makebox[\h][r]{\textsc{Torus}\hspace{18.5mm}\footnotesize{MSE}} \hfill
    \raisebox{21.75mm}[0pt][0pt]{
    \includegraphics[ width=\hh]{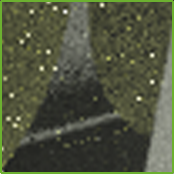}\hfill
    \mbox{\hspace*{0.0mm}}
    \includegraphics[ width=\hh]{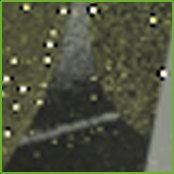}\hfill
    \mbox{\hspace*{0.0mm}}
    \includegraphics[ width=\hh]{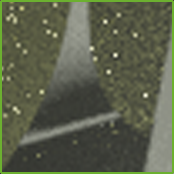}\hfill
    \mbox{\hspace*{0.0mm}}
    \includegraphics[ width=\hh]{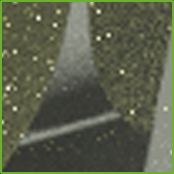}\hfill
    \mbox{\hspace*{0.0mm}}
    \includegraphics[ width=\hh]{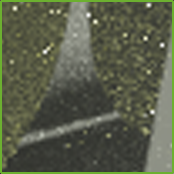}\hfill
    \mbox{\hspace*{0.0mm}}
    \includegraphics[ width=\hh]{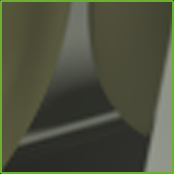}\hfill
    }\\
    \makebox[\h][r]{\footnotesize{1-SSIM}} \hfill
    \raisebox{7.75mm}[0pt][0pt]{
    \includegraphics[ width=\hh]{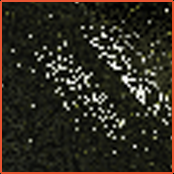}\hfill
    \mbox{\hspace*{0.0mm}}
    \includegraphics[ width=\hh]{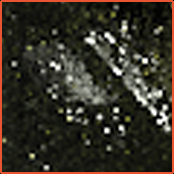}\hfill
    \mbox{\hspace*{0.0mm}}
    \includegraphics[ width=\hh]{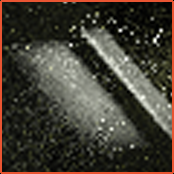}\hfill
    \mbox{\hspace*{0.0mm}}
    \includegraphics[ width=\hh]{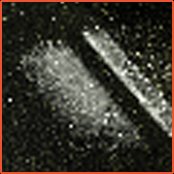}\hfill
    \mbox{\hspace*{0.0mm}}
    \includegraphics[ width=\hh]{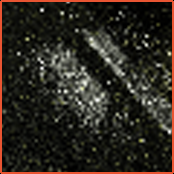}\hfill
    \mbox{\hspace*{0.0mm}}
    \includegraphics[ width=\hh]{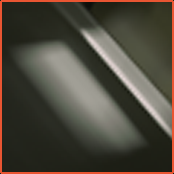}\hfill
    }\\
    \makebox[\h][c]{} \hfill
    \raisebox{8mm}[0pt][0pt]{
    \makebox[\hh][c]{\footnotesize{0.5342}}\hfill
    \mbox{\hspace*{0.0mm}}
    \makebox[\hh][c]{\footnotesize{0.7413}}\hfill
    \mbox{\hspace*{0.0mm}}
    \makebox[\hh][c]{\footnotesize{0.2996}}\hfill
    \mbox{\hspace*{0.0mm}}
    \makebox[\hh][c]{\footnotesize{0.7305}}\hfill
    \mbox{\hspace*{0.0mm}}
    \makebox[\hh][c]{\footnotesize{0.6920}}\hfill
    \mbox{\hspace*{0.0mm}}
    \makebox[\hh][c]{\footnotesize{}}\hfill
    }\\
    \makebox[\h][c]{} \hfill
    \raisebox{8mm}[0pt][0pt]{
    \makebox[\hh][c]{\footnotesize{0.3946}}\hfill
    \mbox{\hspace*{0.0mm}}
    \makebox[\hh][c]{\footnotesize{0.2307}}\hfill
    \mbox{\hspace*{0.0mm}}
    \makebox[\hh][c]{\footnotesize{0.2102}}\hfill
    \mbox{\hspace*{0.0mm}}
    \makebox[\hh][c]{\footnotesize{0.2210}}\hfill
    \mbox{\hspace*{0.0mm}}
    \makebox[\hh][c]{\footnotesize{0.2620}}\hfill
    \mbox{\hspace*{0.0mm}}
    \makebox[\hh][c]{\footnotesize{}}\hfill
    }\\
    \vspace*{-14.5mm}
    \caption{Equal sample count (128 spp) comparisons of path tracing without PSS warping, kd-tree based PSS warping~\shortcite{Guo18}, and our approach. We warp the first one, two, and three path bounce(s), corresponding to 4D, 6D and 8D in PSS. {The training datasets use \textit{epp}-16. Due to the small training data, warping in 6D and 8D does not provide benefits over 4D warping in most scenes.}}
    \label{fig:path_tracing}
\end{figure*}

In Figure~\ref{fig:path_tracing}, we apply our method to render a range of challenging scenes.  \textsc{Country Kitchen} and \textsc{White Room} contain complex visibility settings, and they are dominated by indirect illumination. \textsc{Salle De Bain}, \textsc{Classroom} and \textsc{Torus} feature glossy and specular light transport. Our method treats the underlying Monte Carlo ray tracing method (path tracing) as a black box, and globally warps light transport paths in the PSS.

We compare our method with baseline path tracing (PT without PSS coordinates warping) and a recent kd-tree based PSS warping method~\shortcite{Guo18}. Since our method and other methods differ in the rendering mode of GPU and CPU computations, we show equal sample budget comparisons. Timings for rendering and the ratios of zero-radiance paths are presented in Table~\ref{tab:stats}. Our method can learn the distribution of path contributions and distributes samples according to the distribution, thus it effectively reduces the ratio of zero-radiance paths, which carry no radiance. {Time cost of training neural networks is tabulated in Table~\ref{tab:trainingtime}}. 
We apply two metrics to measure quality of images: mean squared errors (MSE) and Structural Similarity Index (SSIM)~\shortcite{Wang04}. Note that we present the values $1-SSIM$. Hence lower values indicate higher image quality. Finally, all reference images are rendered with at least 32768 samples per pixel (\emph{spp}) to get noise-free ground truth.

\begin{figure*}[tb]
\begin{center}
\begin{tabular*}{0.80\linewidth}{@{}c@{ }c@{ }c@{}}
\scriptsize{Country Kitchen} & \scriptsize{Pool Ball} & \scriptsize{Natural History}\\[-0.3mm]
\begin{minipage}[t]{0.2666\linewidth}
    \centering
    \includegraphics[width=1.0\linewidth]{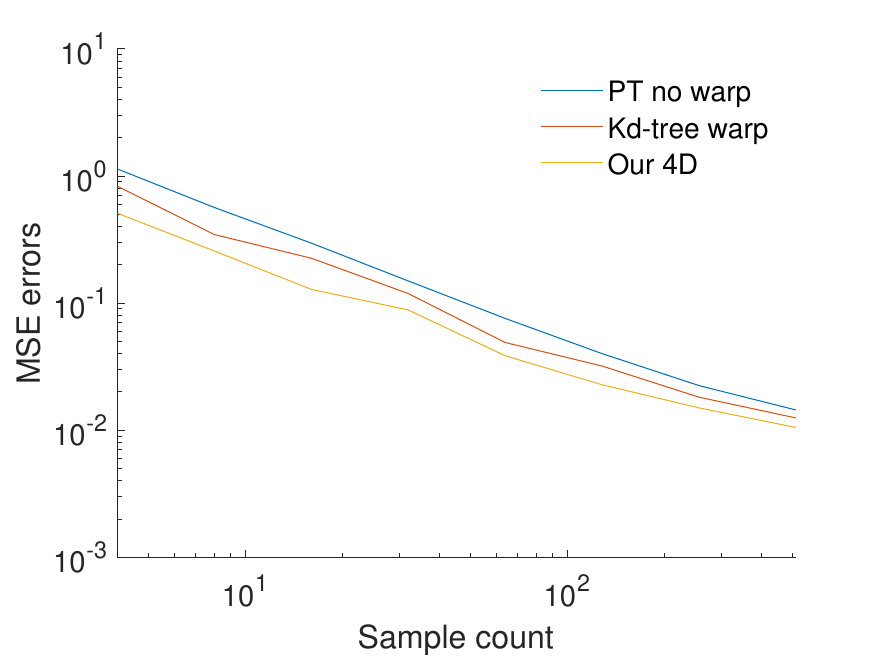}
\end{minipage}&
\begin{minipage}[t]{0.2666\linewidth}
    \centering
    \includegraphics[width=1.0\linewidth]{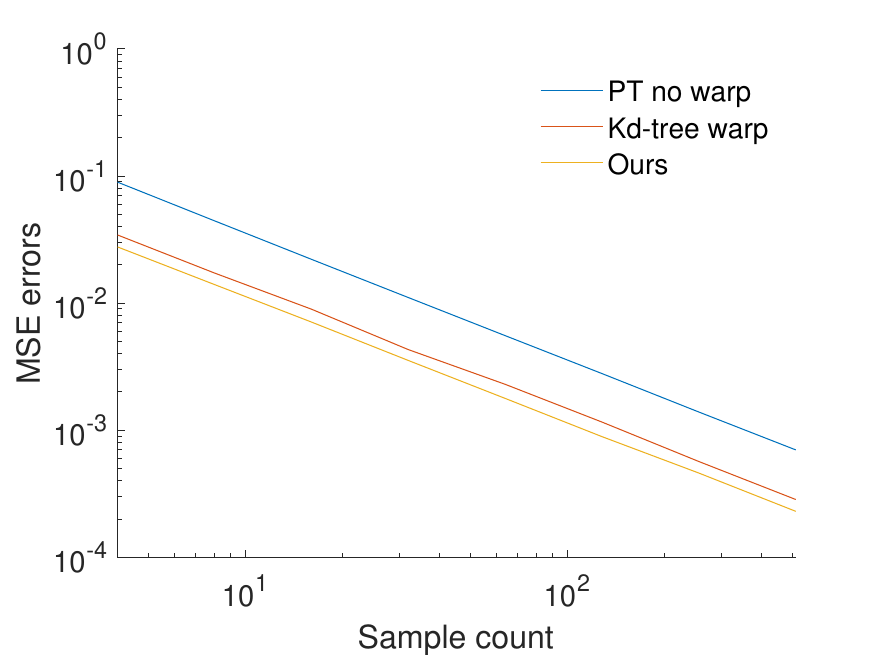}
\end{minipage}&
\begin{minipage}[t]{0.2666\linewidth}
    \centering
    \includegraphics[width=1.0\linewidth]{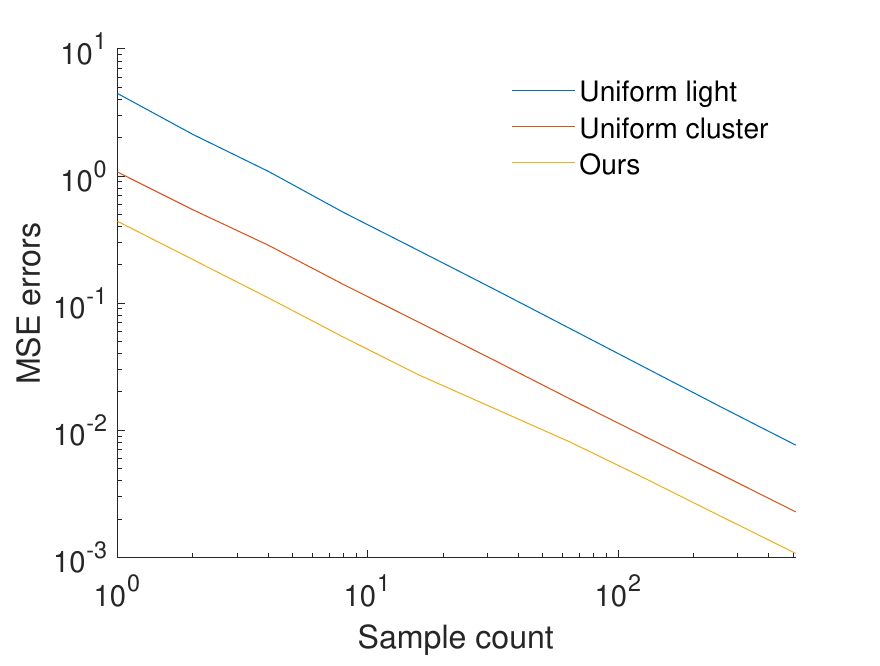}
\end{minipage}\\[-1mm]
\end{tabular*}
\end{center}
\caption{\label{fig:mse_converge}MSE convergence. For the first two scenes, we compare our method against PT without PSS warp and kd-tree based warp. For the \textsc{Natural History} scene, we compare our method against uniformly sampling one light and uniformly sampling one cluster.}
\end{figure*}

\begin{table*}
\centering
\caption{\label{tab:stats}Comparisons of time cost of rendering of different techniques and the ratios of zero-radiance paths (Black ratio). For all method, images are rendered with 16 \emph{spp} using 2 threads. Timings for our method include both the sample generation and rendering process.}
\setlength{\tabcolsep}{6pt}
\renewcommand{\arraystretch}{1}
\begin{tabular*}{1.0\textwidth}{ccccccccccc}
\toprule
\multirow{2}{*}{Scene}& \multicolumn{2}{c}{PT no warp}&\multicolumn{2}{c}{kd-tree warp}&\multicolumn{2}{c}{Our 4D}&\multicolumn{2}{c}{Our 6D}&\multicolumn{2}{c}{Our 8D}\\
\cmidrule(lr){2-3} \cmidrule(lr){4-5} \cmidrule(lr){6-7} \cmidrule(lr){8-9} \cmidrule(lr){10-11}
{} & Timing & Blk. ratio & Timing & Blk. ratio & Timing & Blk. ratio & Timing & Blk. ratio & Timing & Blk. ratio\\
\midrule 
{Country Kitchen} & 65s & {\makebox[10mm][r]{$27.16\%$}} & 136s & {\makebox[10mm][r]{$18.39\%$}} & 170.8s & {\makebox[10mm][r]{$9.26\%$}} & 206.3s & {\makebox[10mm][r]{$9.54\%$}} & 238.6s & {\makebox[10mm][r]{$9.62\%$}}\\
{Salle De Bain}   & 54s & {\makebox[10mm][r]{$8.94\%$}} & 121s & {\makebox[10mm][r]{$24.29\%$}} & 177.4s & {\makebox[10mm][r]{$4.70\%$}} & 186.2s & {\makebox[10mm][r]{$4.45\%$}} & 221.5s & {\makebox[10mm][r]{$6.92\%$}}\\
{White Room}      & 69s & {\makebox[10mm][r]{$18.17\%$}} & 162s & {\makebox[10mm][r]{$11.91\%$}} & 184.3s & {\makebox[10mm][r]{$8.61\%$}} & 231.1s & {\makebox[10mm][r]{$7.79\%$}} & 272.2s & {\makebox[10mm][r]{$9.30\%$}}\\
{Classroom}       & 61s & {\makebox[10mm][r]{$69.79\%$}} & 98s & {\makebox[10mm][r]{$62.74\%$}} & 210.8s & {\makebox[10mm][r]{$59.60\%$}} & 240.4s & {\makebox[10mm][r]{$60.29\%$}} & 274.6s & {\makebox[10mm][r]{$60.67\%$}}\\
{Torus}           & 14s &  $19.86\%$ & 42s & $42.65\%$ & 154.5s & $13.28\%$ & 168.1s & {\makebox[10mm][r]{$20.60\%$}} & 208.2s & $14.45\%$\\
\bottomrule
\end{tabular*}
\end{table*}

\begin{table}
\centering
\caption{\label{tab:trainingtime}{Time cost of training neural networks for 4D, 6D and 8D data in minutes (m). For each scene, the training example count is set to epp times the image resolution. Neural networks are trained for 60 epochs on an Nvidia GeForce GTX 1070 GPU.}}
\setlength{\tabcolsep}{3pt}
\begin{tabular*}{0.9\linewidth}{rcccc}
\toprule
Scene & Example\# & {4D} & {6D} & {8D}\\
\midrule
{\small{Country Kitchen}} & \small{$16\times160\times88$} & {\makebox[8mm][r]{\small{19.4m}}} & {\makebox[8mm][r]{\small{20.2m}}} & {\makebox[8mm][r]{\small{20.8m}}}\\
{\small{Salle De Bain}} & \small{$16\times160\times88$} & {\makebox[8mm][r]{\small{11.6m}}} & {\makebox[8mm][r]{\small{12.1m}}} & {\makebox[8mm][r]{\small{12.3m}}}\\
{\small{White Room}} & \small{$16\times160\times88$} & {\makebox[8mm][r]{\small{15.0m}}} & {\makebox[8mm][r]{\small{15.6m}}} & {\makebox[8mm][r]{\small{15.9m}}}\\
{\small{Classroom}} & \small{$16\times160\times88$} & {\makebox[8mm][r]{\small{17.0m}}} & {\makebox[8mm][r]{\small{17.2m}}} & {\makebox[8mm][r]{\small{17.5m}}}\\
{\small{Torus}} & \small{$16\times100\times100$} & {\makebox[8mm][r]{\small{9.2m}}} & {\makebox[8mm][r]{\small{9.4m}}} & {\makebox[8mm][r]{\small{10.7m}}}\\
\bottomrule
\end{tabular*}
\end{table}

As described in Section~\ref{sec:rendering}, we apply PSS warping to the path prefixes of one, two or three bounce(s), corresponding to dimensionalities of 4D, 6D and 8D in the primary sample space. In our experiments, {next event estimation which is part of the black box, is enabled for path tracing.} {For the kd-tree based method~\shortcite{Guo18}, we use its default settings}.

As can be observed in Figure~\ref{fig:path_tracing}, our method in 4D and 6D, corresponding to warping the first one or two bounce(s), generally provides lower errors than the baselines. A similar finding is reported in~\shortcite{Guo18}, where they only warp the initial two bounces. Interestingly, warping more than six dimensions leads to slightly higher errors. This is mainly because the degrees of freedom of a PSS vector increase with the number of dimensions. {It leads to overfitting issues that could be avoided by providing more training data, at the cost of additional training time. Additionally, since paths with more vertices typically make smaller contributions to the image than paths with fewer vertices, improved importance sampling of vertices further along a path brings decreasing benefits}. A failure case of our method is the \textsc{Classroom} scene, where our method does not reduce errors. This scene is equipped with many small-scale geometries, such as desks and chairs with glossy legs, which can undermine the correspondence between well-distributed PSS vectors and light transport paths.

\begin{figure}[H]
\vspace{-1.0em}
\begin{center}
\begin{tabular}{@{}c@{}c@{ }c@{ }c@{}}
\footnotesize{Reference} & \footnotesize{PT no warp} & \footnotesize{Kd-tree warp} & \footnotesize{Ours}\\
\multirow{2}{*}[13.35mm]{
\begin{minipage}[t]{0.4\linewidth}
    \centering
    \includegraphics[width=1.0\linewidth]{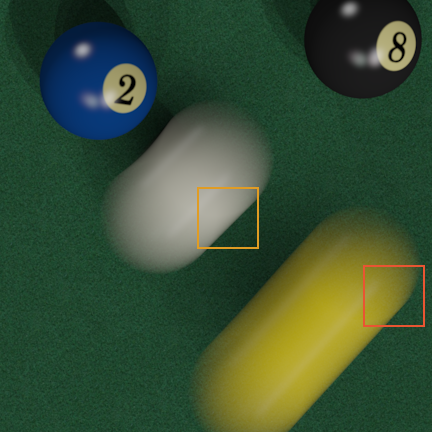}
\end{minipage}
}&
\begin{minipage}[t]{0.190\linewidth}
    \centering
    \includegraphics[width=1.0\linewidth]{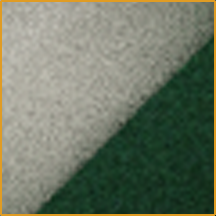}
\end{minipage}&
\begin{minipage}[t]{0.190\linewidth}
    \centering
    \includegraphics[width=1.0\linewidth]{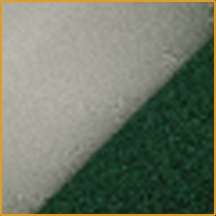}
\end{minipage}&
\begin{minipage}[t]{0.190\linewidth}
    \centering
    \includegraphics[width=1.0\linewidth]{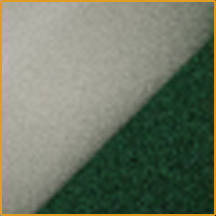}
\end{minipage}\\
 &
\begin{minipage}[t]{0.190\linewidth}
    \centering
    \includegraphics[width=1.0\linewidth]{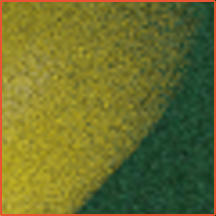}
\end{minipage}&
\begin{minipage}[t]{0.190\linewidth}
    \centering
    \includegraphics[width=1.0\linewidth]{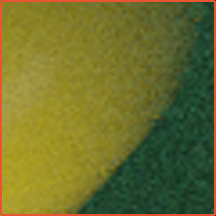}
\end{minipage}&
\begin{minipage}[t]{0.190\linewidth}
    \centering
    \includegraphics[width=1.0\linewidth]{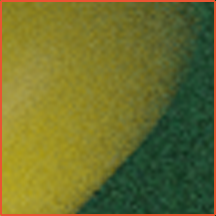}
\end{minipage}\\[-1mm]
\makebox[0.4\linewidth][r]{\footnotesize{MSE}}&
\footnotesize{0.00278} & \footnotesize{0.00104} & \footnotesize{0.00089}\\[-1mm]
\makebox[0.4\linewidth][r]{\footnotesize{1-SSIM}}&
\footnotesize{0.29600} & \footnotesize{0.23210} & \footnotesize{0.22370}\\
\multirow{2}{*}[13.35mm]{
\begin{minipage}[t]{0.4\linewidth}
    \centering
    \includegraphics[width=1.0\linewidth]{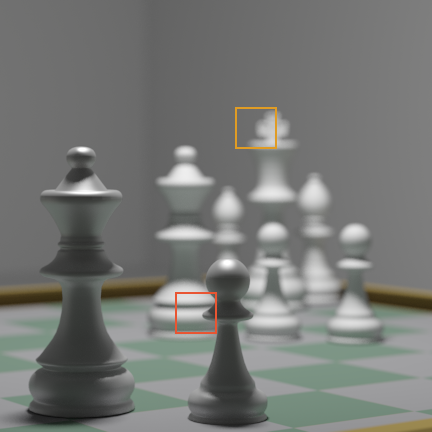}
\end{minipage}
}&
\begin{minipage}[t]{0.190\linewidth}
    \centering
    \includegraphics[width=1.0\linewidth]{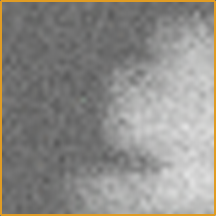}
\end{minipage}&
\begin{minipage}[t]{0.190\linewidth}
    \centering
    \includegraphics[width=1.0\linewidth]{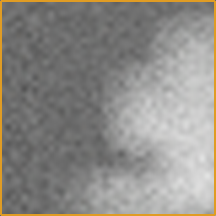}
\end{minipage}&
\begin{minipage}[t]{0.190\linewidth}
    \centering
    \includegraphics[width=1.0\linewidth]{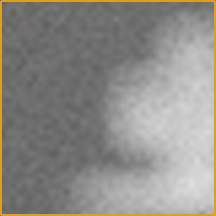}
\end{minipage}\\
 &
\begin{minipage}[t]{0.190\linewidth}
    \centering
    \includegraphics[width=1.0\linewidth]{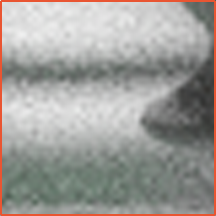}
\end{minipage}&
\begin{minipage}[t]{0.190\linewidth}
    \centering
    \includegraphics[width=1.0\linewidth]{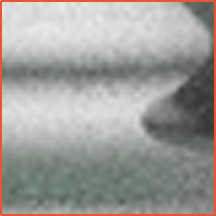}
\end{minipage}&
\begin{minipage}[t]{0.190\linewidth}
    \centering
    \includegraphics[width=1.0\linewidth]{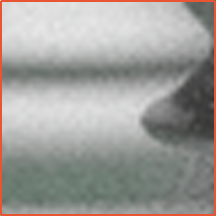}
\end{minipage}\\[-1mm]
\makebox[0.4\linewidth][r]{\footnotesize{MSE}}&
\footnotesize{0.00944} & \footnotesize{0.00721} & \footnotesize{0.00346}\\[-1mm]
\makebox[0.4\linewidth][r]{\footnotesize{1-SSIM}}&
\footnotesize{0.32990} & \footnotesize{0.24950} & \footnotesize{0.24620}\\
\end{tabular}
\end{center}
\vspace{-0.5em}
\caption{\label{fig:distrib_effect}Comparisons of distribution effects: motion blur (Top: \textsc{Pool Ball}) and depth of field (Bottom: \textsc{Chess}). {Training data set here uses \textit{epp}-16.}}
\vspace{-1em}
\end{figure}

In Figure~\ref{fig:distrib_effect}, we demonstrate the ability of our method to render distribution effects, such as motion blur (\textsc{Pool Ball} scene) and defocus blur (\textsc{Chess} scene). Motion blur is controlled by time samples and shutter opening intervals, and defocus blur is affected by lens samples and aperture radius. Therefore, we focus on warping only the first 3D or 4D of a PSS vector, corresponding to components of camera samples, and use uniform random data for the remaining dimensions. Compared to the baseline approach without PSS warping and kd-tree based warping, our method achieves lower numerical errors and better image quality.

\begin{figure}[H]
\begin{center}
\begin{tabular}{@{}c@{}c@{}c@{}}
\begin{minipage}[c]{0.33\linewidth}
    \centering
    \includegraphics[width=1.0\linewidth]{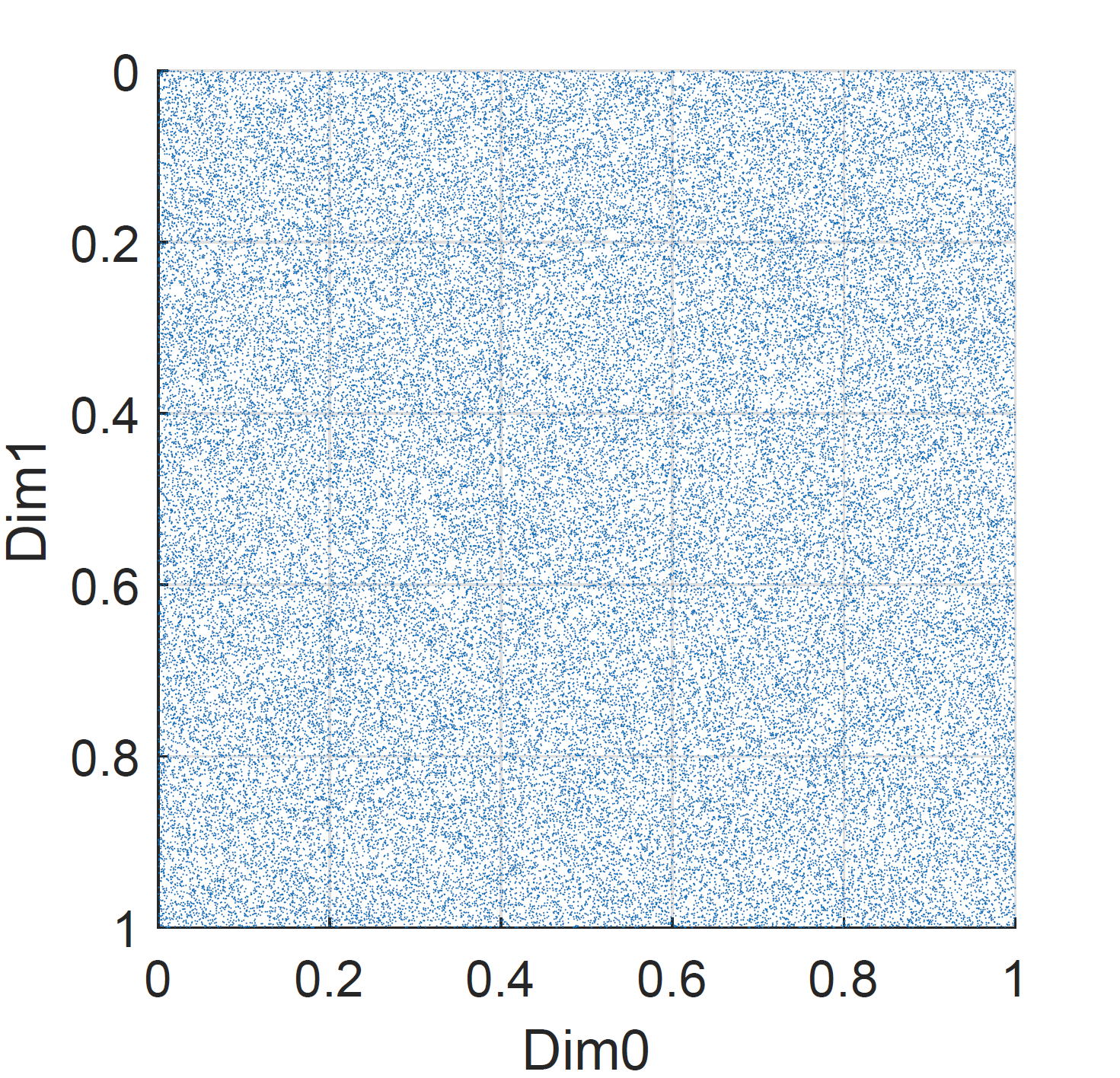}
\end{minipage}&
\begin{minipage}[c]{0.33\linewidth}
    \centering
    \includegraphics[width=1.0\linewidth]{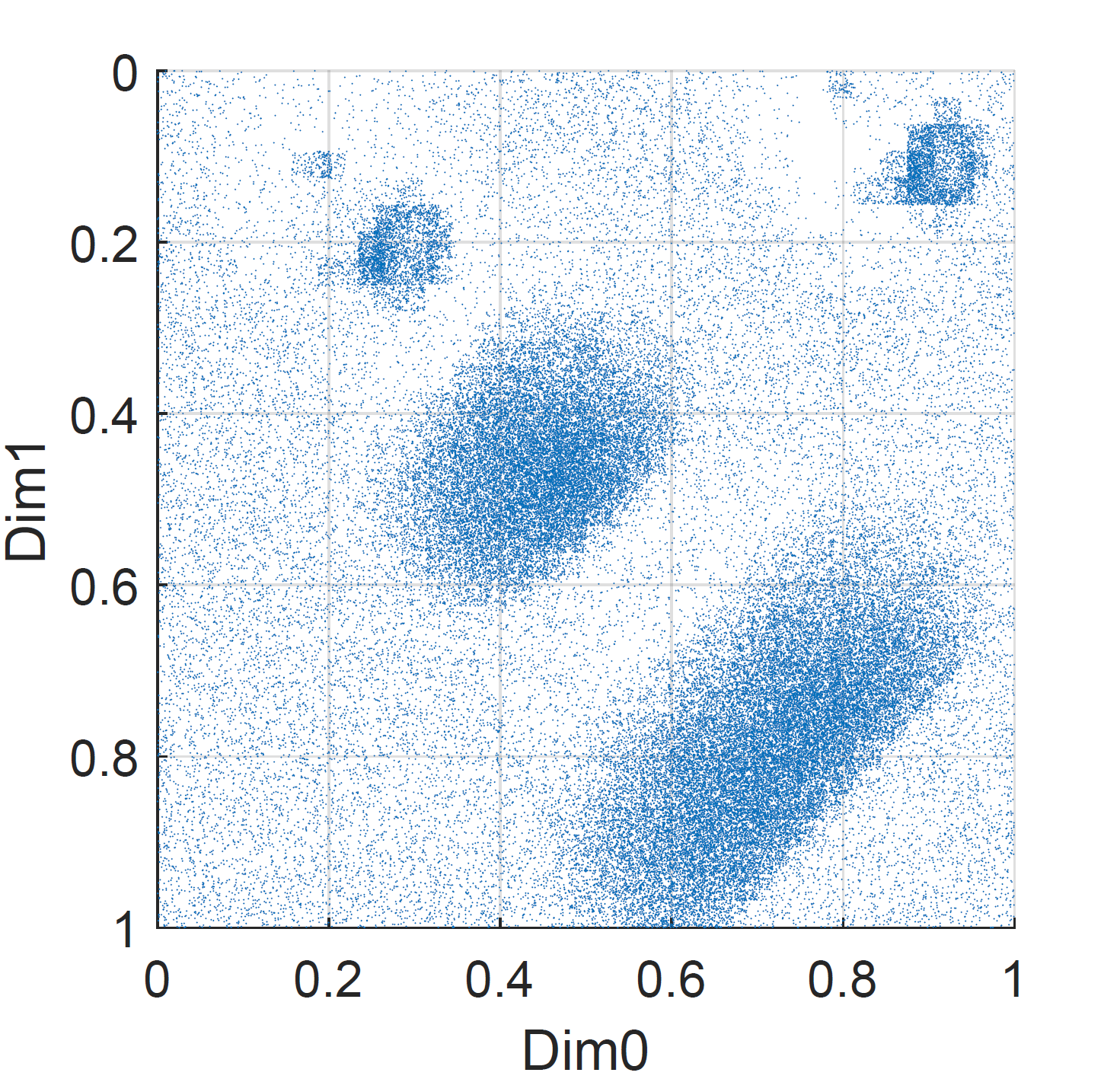}
\end{minipage}&
\begin{minipage}[c]{0.33\linewidth}
    \centering
    \includegraphics[width=1.0\linewidth]{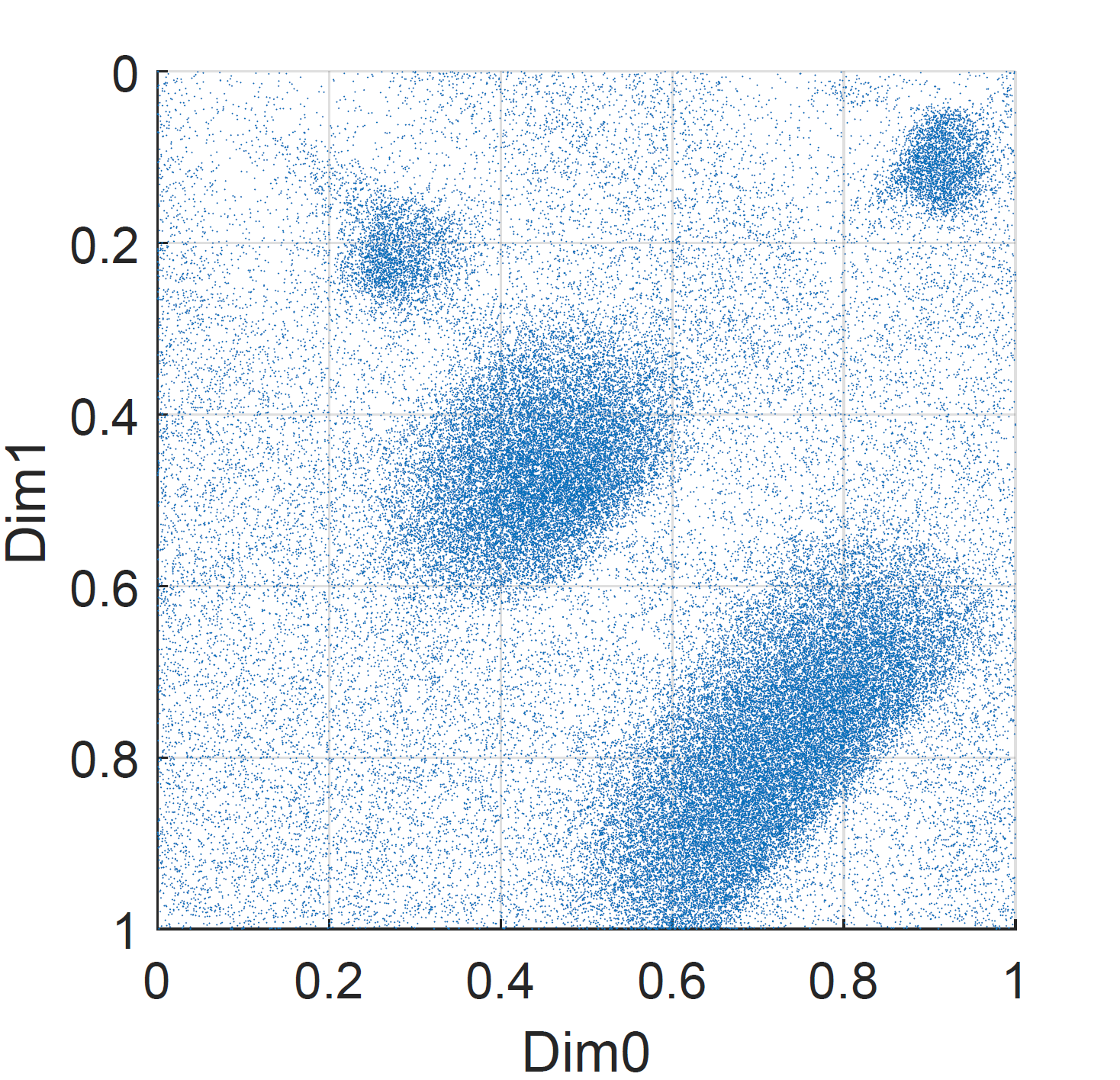}
\end{minipage}\\
\begin{minipage}[c]{0.33\linewidth}
    \centering
    \includegraphics[width=1.0\linewidth]{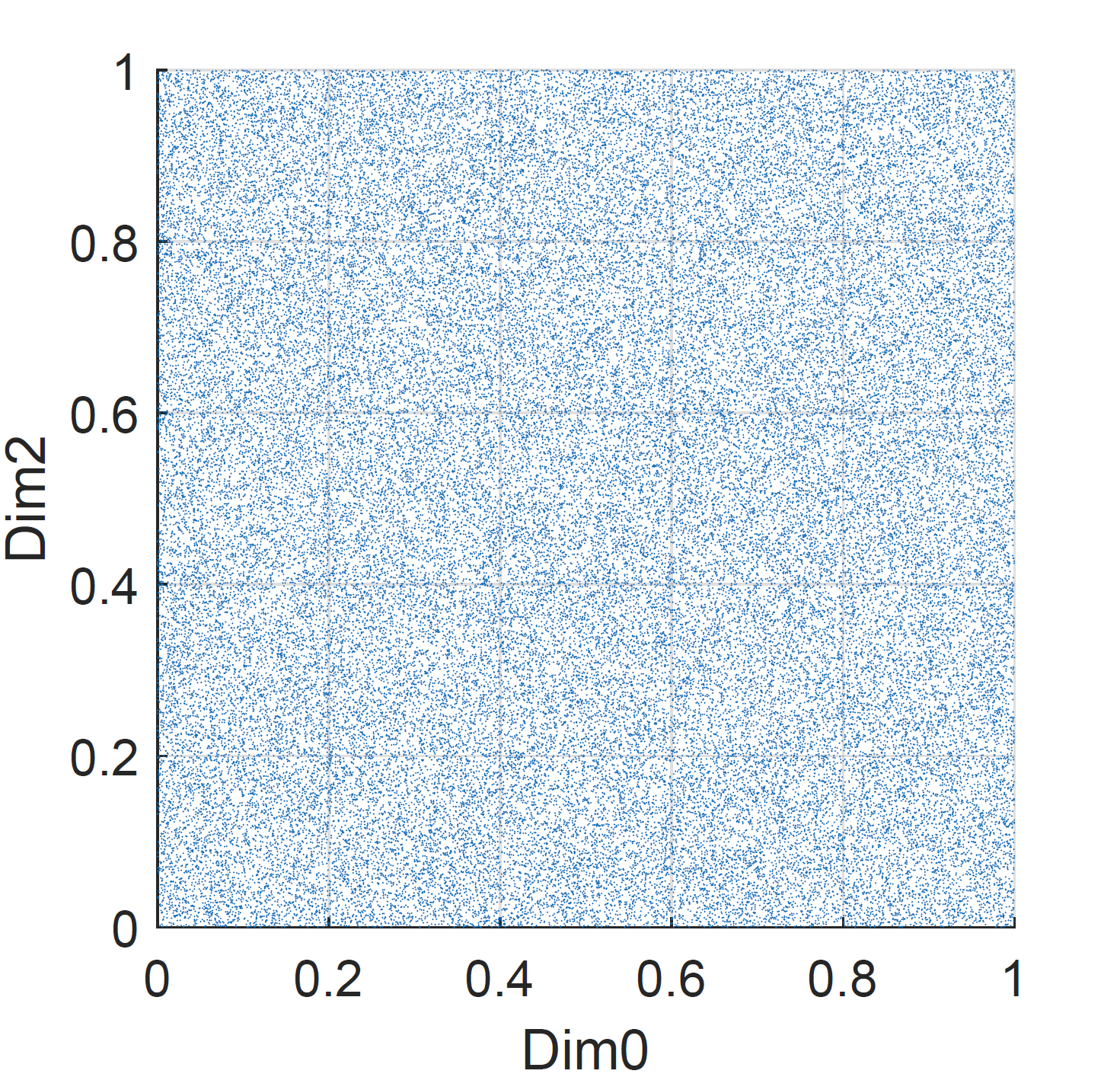}
\end{minipage}&
\begin{minipage}[c]{0.33\linewidth}
    \centering
    \includegraphics[width=1.0\linewidth]{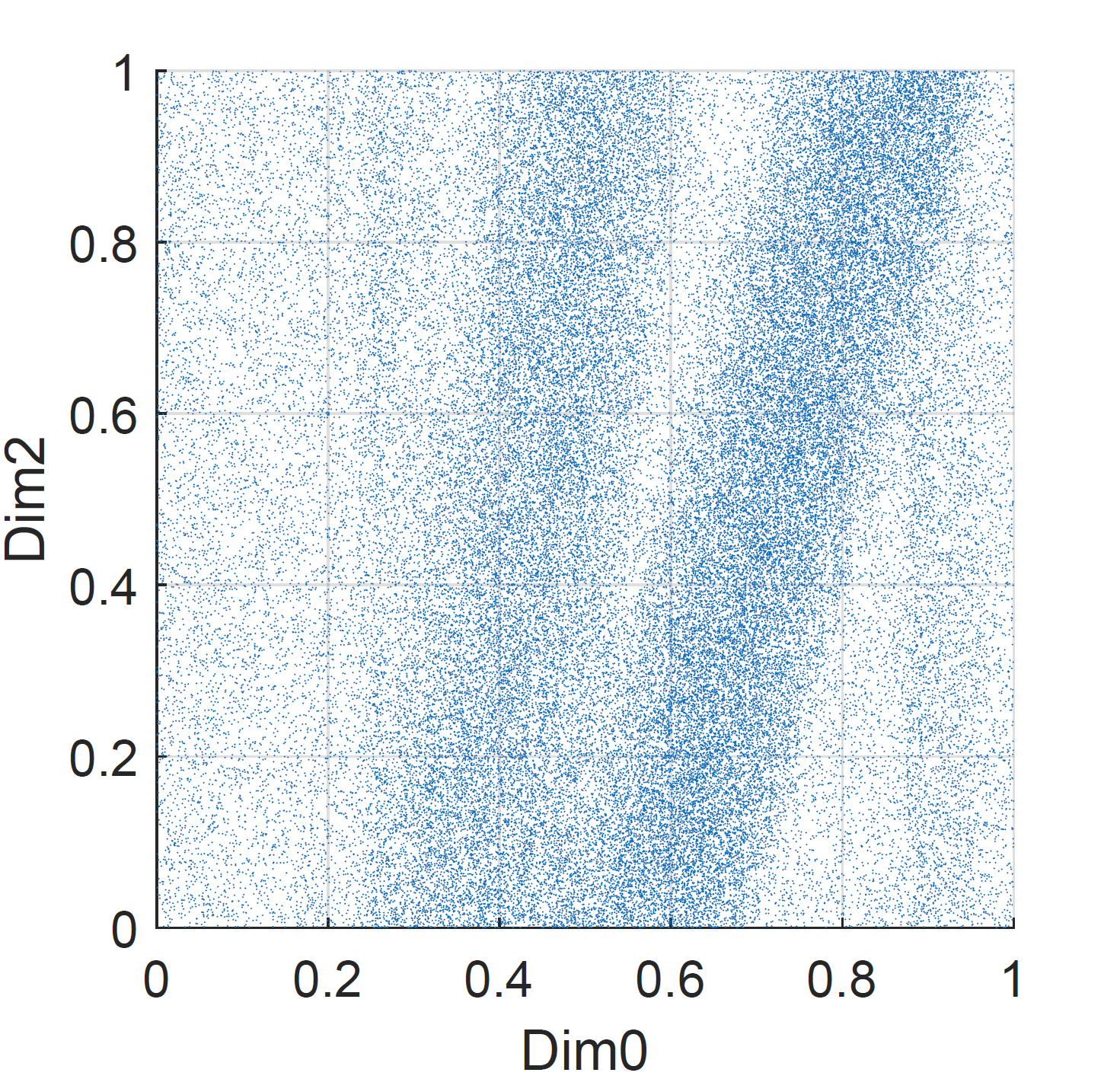}
\end{minipage}&
\begin{minipage}[c]{0.33\linewidth}
    \centering
    \includegraphics[width=1.0\linewidth]{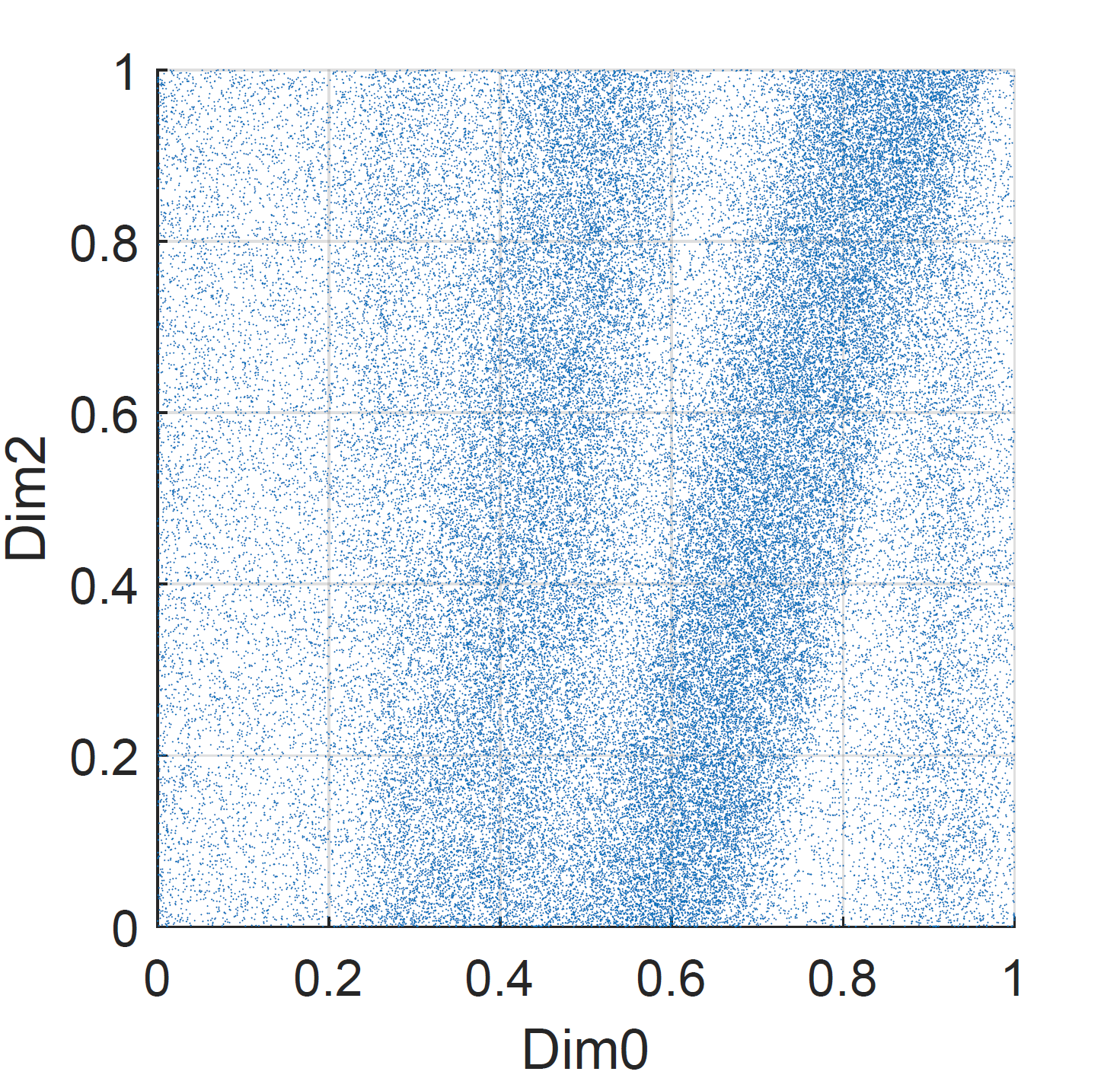}
\end{minipage}\\
\begin{minipage}[c]{0.33\linewidth}
    \centering
    \includegraphics[width=1.0\linewidth]{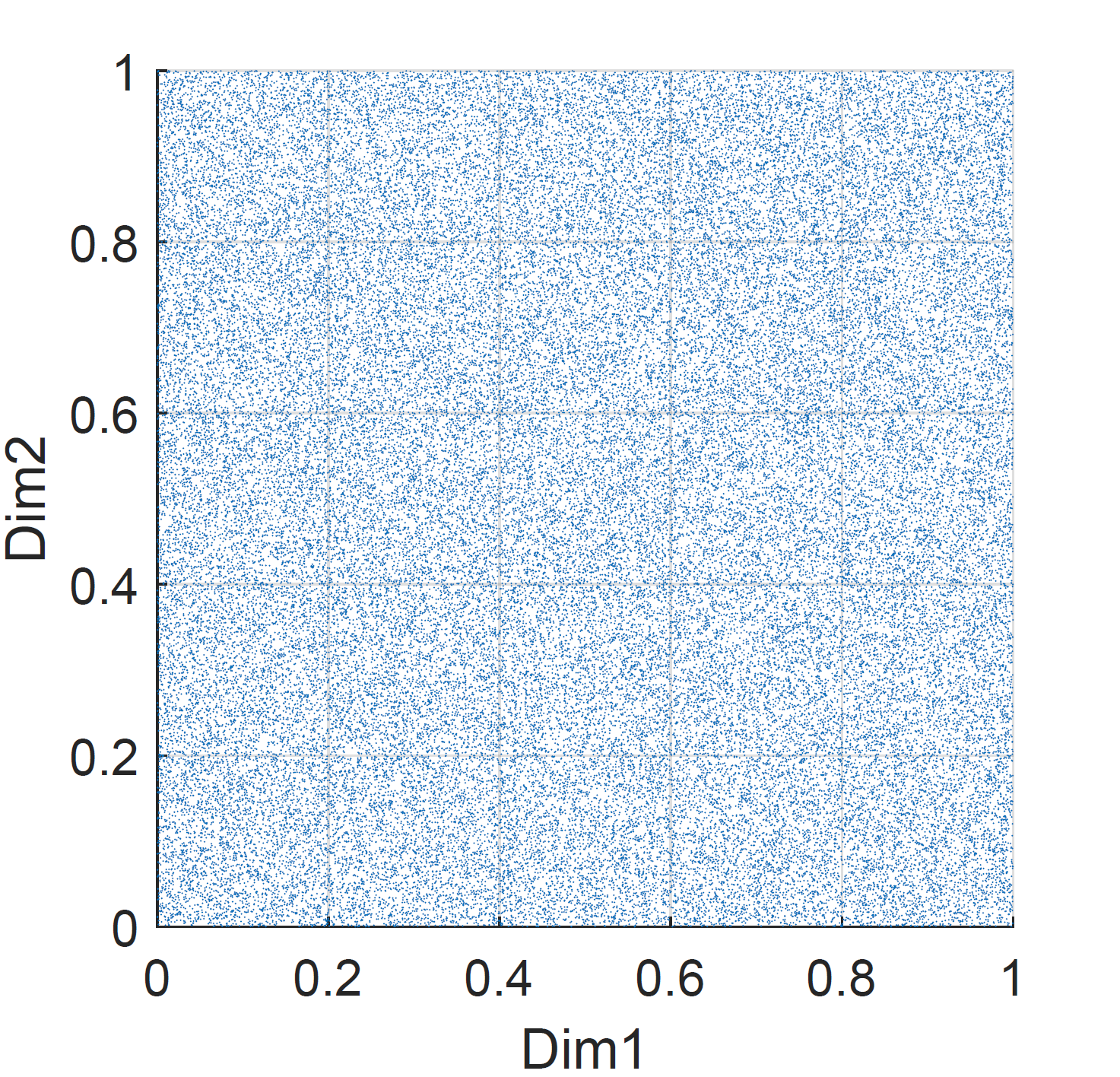}
\end{minipage}&
\begin{minipage}[c]{0.33\linewidth}
    \centering
    \includegraphics[width=1.0\linewidth]{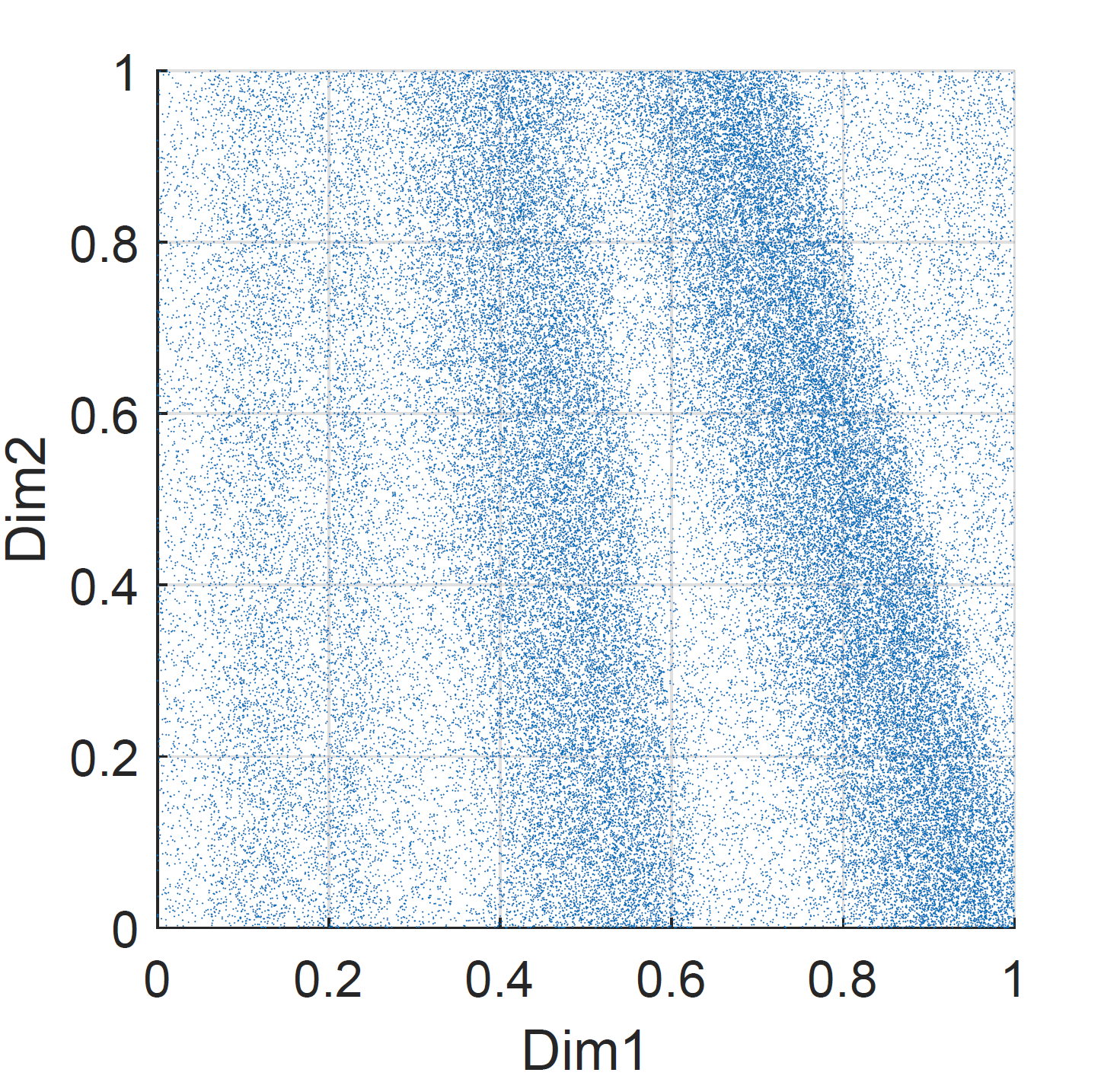}
\end{minipage}&
\begin{minipage}[c]{0.33\linewidth}
    \centering
    \includegraphics[width=1.0\linewidth]{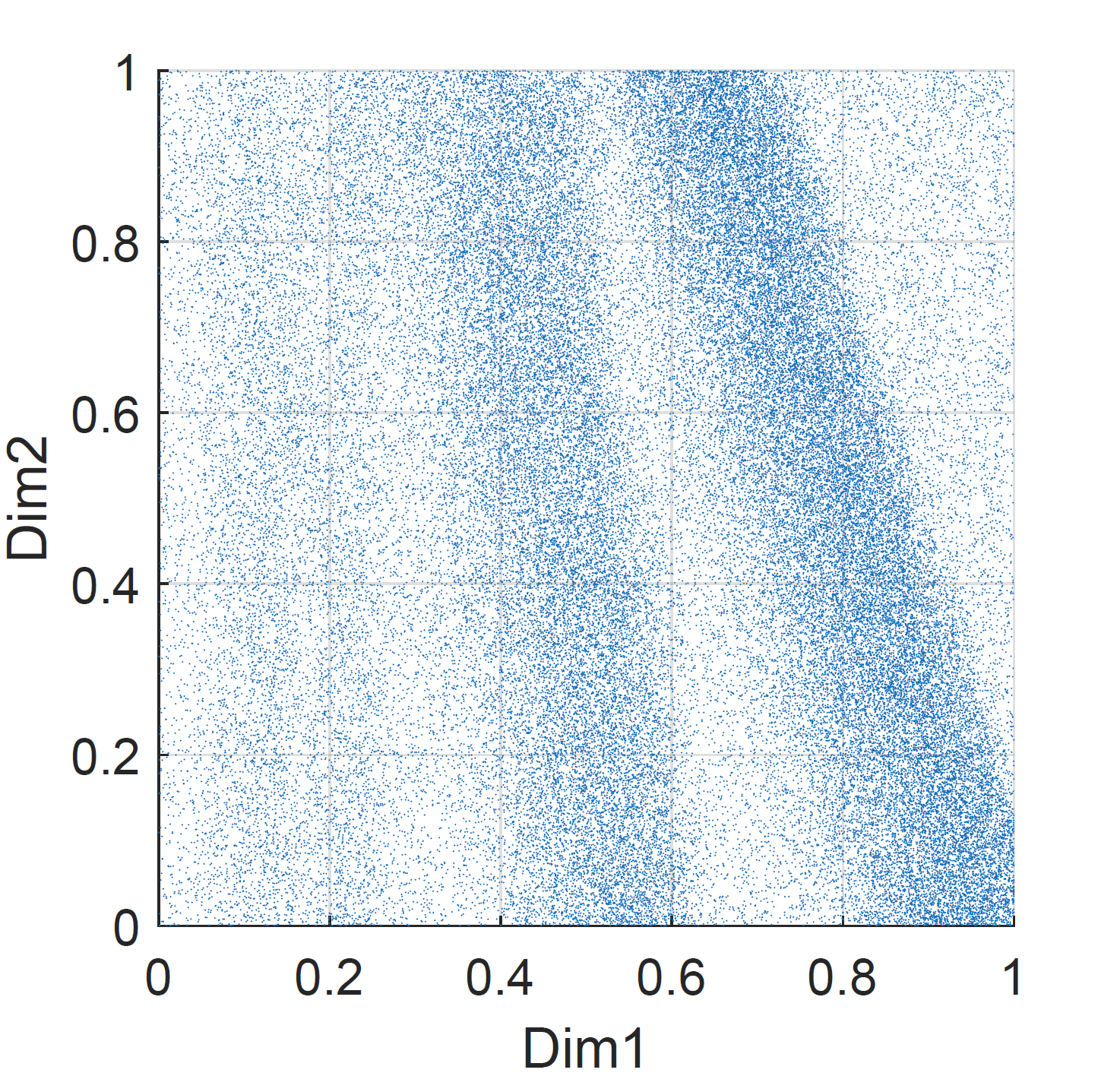}
\end{minipage}\\
{\small{(a)}}&{\small(b)}&{\small(c)}\\
\end{tabular}
\end{center}
\caption{\label{fig:sample_distribution}Visualizations of sample distributions of the \textsc{Pool Ball} scene's 3D warping. We compare distributions of $80,000$ uniform random input samples (a), warped samples of kd-tree based approach (b) and our method (c). Our method distributes more samples to domain corresponding to motion blur and better cover the whole sampling domain, while the sample distribution of kd-tree warping tends to be blocky. The Dim denotes the dimension.}
\end{figure}

To obtain an interpretable profile of a warped distribution, we further visualize distributions of uniform random input samples and warped samples of the \textsc{Pool Ball} scene in Figure~\ref{fig:sample_distribution}. In this case, 3D PSS warping provides us with an intuitively meaningful sampling distribution. As can be seen, our method is able to warp samples to increase the density in the area corresponding to motion blur, while covering the rest of the domain with lower density. {In contrast}, the sample distribution of kd-tree warping tends to be blocky.

\begin{figure}
\begin{center}
\setlength{\tabcolsep}{0.0pt}
\begin{tabular}{@{}c@{}c@{ }c@{ }c@{}}
\footnotesize{Reference} & \footnotesize{(a)} & \footnotesize{(b)} & \footnotesize{(c)}\\
\multirow{2}{*}[12.20mm]{
\begin{minipage}[t]{0.4520\linewidth}
    \centering
    \includegraphics[width=1.0\linewidth]{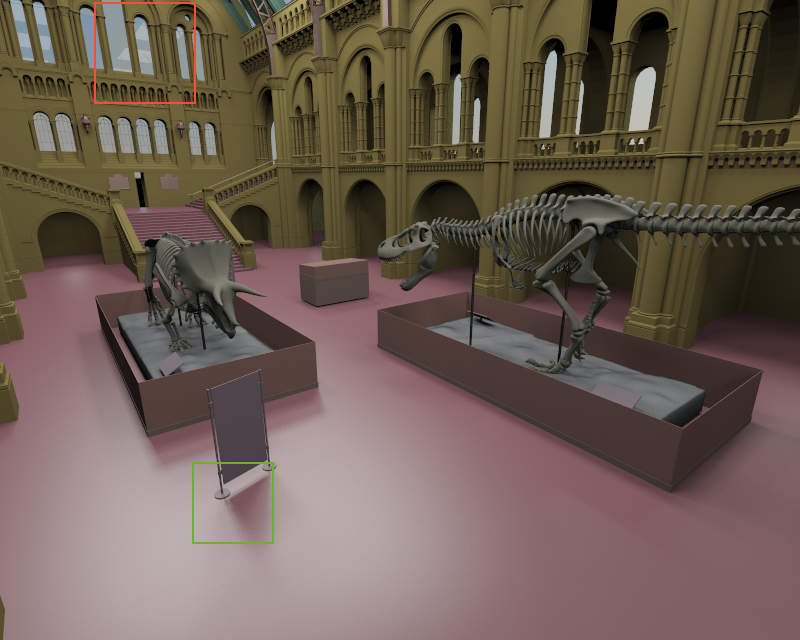}
\end{minipage}
}&
\begin{minipage}[t]{0.1770\linewidth}
    \centering
    \includegraphics[width=1.0\linewidth]{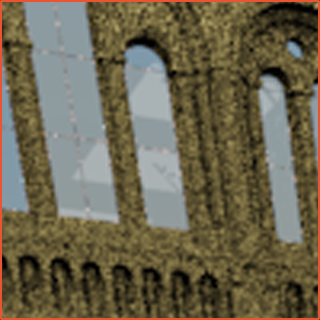}
\end{minipage}&
\begin{minipage}[t]{0.1770\linewidth}
    \centering
    \includegraphics[width=1.0\linewidth]{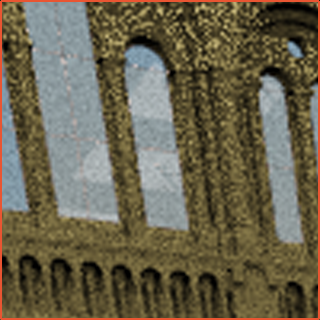}
\end{minipage}&
\begin{minipage}[t]{0.1770\linewidth}
    \centering
    \includegraphics[width=1.0\linewidth]{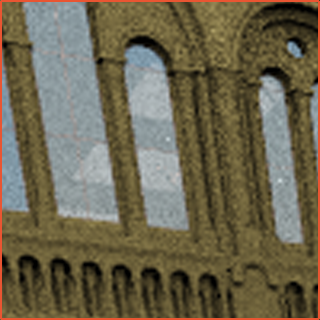}
\end{minipage}\\[-0.3mm]
 &
\begin{minipage}[t]{0.1770\linewidth}
    \centering
    \includegraphics[width=1.0\linewidth]{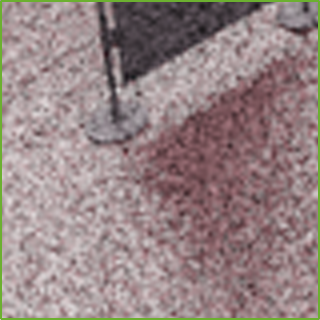}
\end{minipage}&
\begin{minipage}[t]{0.1770\linewidth}
    \centering
    \includegraphics[width=1.0\linewidth]{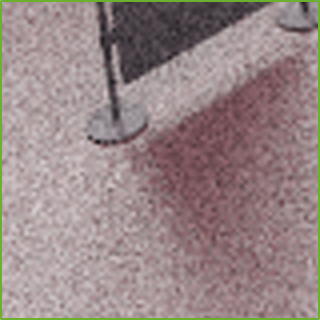}
\end{minipage}&
\begin{minipage}[t]{0.1770\linewidth}
    \centering
    \includegraphics[width=1.0\linewidth]{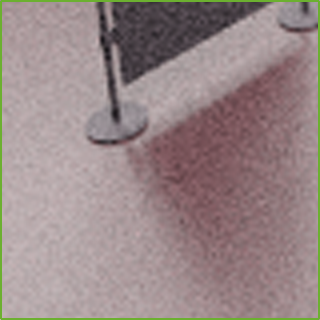}
\end{minipage}\\
\makebox[0.4460\linewidth][r]{\footnotesize{MSE}}&
\footnotesize{0.06328} & \footnotesize{0.01767} & \footnotesize{0.01039}\\[-1mm]
\makebox[0.4460\linewidth][r]{\footnotesize{1-SSIM}}&
\footnotesize{0.3755} & \footnotesize{0.3170} & \footnotesize{0.2218}\\
\end{tabular}
\end{center}
\caption{\label{fig:directlighting}Visual comparisons for complex many-light rendering. We compare uniformly sampling one light (a), uniformly sampling one cluster (b) with our method (c), under the the same sample budget of 64 spp. {The training dataset uses \textit{epp}-16.}}
\end{figure}

In addition, our method can be applied to render complex many-light direct illumination. The \textsc{Natural History} scene in Figure~\ref{fig:directlighting} contains $92$ area light sources and one environment light map, posing a challenge to traditional light selection and accumulation techniques. We firstly build a light tree to organize all light sources and split the scene's space into multiple disjoint regions. Then we compute light clusters (a.k.a., a lightcut~\shortcite{Walter:2005:LSA}) for each region. Subsequently, {we importance sample a 3D PSS distribution to get coordinates for selecting components with high contribution}. We compare our method with the baseline method of uniformly sampling each individual light and uniformly sampling each cluster of a lightcut. The baseline method is unaware of the visibility difficulty, thus most shadow rays are blocked and computation efforts are wasted. Our method can learn the shape of the manifold of light path samples in the PSS and generate new samples on the manifold, thus it achieves a significant error reduction.

Figure~\ref{fig:mse_converge} illustrates MSE errors with respect to the average \emph{spp}. We select one scene from each of the previous three application categories: PSS path warping, distribution effects, and many-light rendering. As can be seen, our method generally provides a consistently lower error and converges without introducing bias.

\section{{Discussion and Limitations}} \label{sec:discussion}

\textbf{Computation cost.} \,A downside of our approach is that it requires a training phase that takes on the order of minutes for typical warp dimensions up to eight. In addition, the accuracy of the density produced by our warp depends on the amount of training data, and using more training data increases training time. During rendering, the evaluation of neural networks to generate samples also incurs computational overhead. Fortunately, limiting the number of warping dimensions leads to a useful trade-off between training time and variance reduction. 

\textbf{Network weights reuse  and retraining.} \,Being a data-driven approach, our training correlates with the current camera view of a scene. Provided a change of camera view, we can reuse weights of a trained neural network of the previous camera view to initialize retraining. The reuse of network weights implicitly exploits the data coherence of different views of a scene, and it significantly accelerates retraining compared to training from scratch; {see Section 1 of the supplementary material for experimental results}.

{\textbf{Higher dimensional warps.} \,We also experimented with higher-dimensional warps up to 12D (see the supplementary document), and were able to obtain error reductions compared to baseline path tracing up to 12D. This requires more training data and time, however, and we generally did not observe any further error reduction compared to 4D or 6D warps. Hence taking advantage of higher dimensional warps in a practical scenario is an interesting avenue for future research.}

{\textbf{Efficiency of resampling.} \,The resampling process includes a step to draw initial examples to form a candidate set and another step to resample examples from the set. Both steps have linear complexity in terms of the number of examples. Since the original distribution is represented by discrete examples in the candidate set, an accurate representation requires a relatively large candidate set and more processing time. Our further experiments (see Section 3 of the supplementary material) investigate the candidate set sizes in terms of a trade-off between efficiency and quality of resampling.}

{\textbf{Small illumination features.} \,Small-scale caustics with fine structures and glints, caused by specular or highly glossy light transport, are generally hard to render for many methods. Path tracing based on uniform PSS sampling may ``miss'' these small features, since it is inefficient to sample light paths for such features. In the supplementary material, we compare our method against baseline path tracing to show that our approach can efficiently sample small illumination features given enough training data.} 

\textbf{Comparison to caching based approaches.} \,In addition to neural network approaches, previous tabulation based methods applied caching structures like histogram~\shortcite{Dahm17}, octree, quadtree~\shortcite{muller2017practical} or local Gaussian-mixture models~\shortcite{vorba2014line} to learn local, directional sampling densities in an \emph{a posterior} manner. These structures are embedded in the path space and they conduct low-dimensional warping in local regions. In contrast, our method works in a PSS space to learn a sampling distribution, and it treats the underlying rendering algorithm as a black box. This makes it more general to be applied to multiple tasks of importance sampling, such as light path sampling, distribution effects' multi-dimensional sampling, and many-light sampling as shown in the previous section.

\section{{Conclusions and Future Work}} \label{sec:conclusion}

We introduced a novel approach to learn importance sampling of entire light paths in primary sample space using a suitable deep neural network architecture. We leverage the neural network to perform a non-linear warp in primary sample space, achieving a desired target density that can further reduce the variance of an existing rendering algorithm, which is treated as a black box by our method. Our experiments demonstrated that this approach can effectively reduce variance in practical scenarios without introducing bias. A main advantage of our approach is that it is agnostic of specific light transport effects in any scene, and the underlying renderer. Therefore, it is easy to implement on top of existing systems. For future work, reducing the computation costs of our approach is an important and meaningful direction. {Another} interesting avenue for future research is to extend our maximum likelihood approach to a Bayesian framework with a prior, which could allow more effective scene-dependent training using fewer samples.

\section*{Acknowledgements}

We thank Jay-Artist, Nacimus, NovaZeeke and Benedikt Bitterli for constructing and distributing the \textsc{Country Kitchen}, \textsc{White Room}, \textsc{Salle De Bain} and \textsc{Classroom} scenes, and T. Hachisuka for the \textsc{Torus} scene. This work was supported by Swiss National Science Foundation project number 169839.

\bibliographystyle{eg-alpha-doi}
\bibliography{bibliography}

\end{document}